\documentclass{article}

\PassOptionsToPackage{numbers, compress}{natbib} 

\usepackage[main, final]{neurips_2026}
\usepackage{graphicx} 
\usepackage{amsmath}
\usepackage{caption}

\usepackage[utf8]{inputenc} 
\usepackage[T1]{fontenc}    
\usepackage{hyperref}       
\usepackage{url}            
\usepackage{booktabs}       
\usepackage{amsfonts}       
\usepackage{nicefrac}       
\usepackage{microtype}      
\usepackage{xcolor}         
\usepackage{caption}
\usepackage[most]{tcolorbox}
\usepackage{multirow}
\usepackage{stfloats}
\usepackage{float}
\usepackage{placeins}
\usepackage{wrapfig}
\usepackage{fontawesome5}   
\newcommand{\hflogo}{\raisebox{-0.15em}{\includegraphics[height=1.05em]{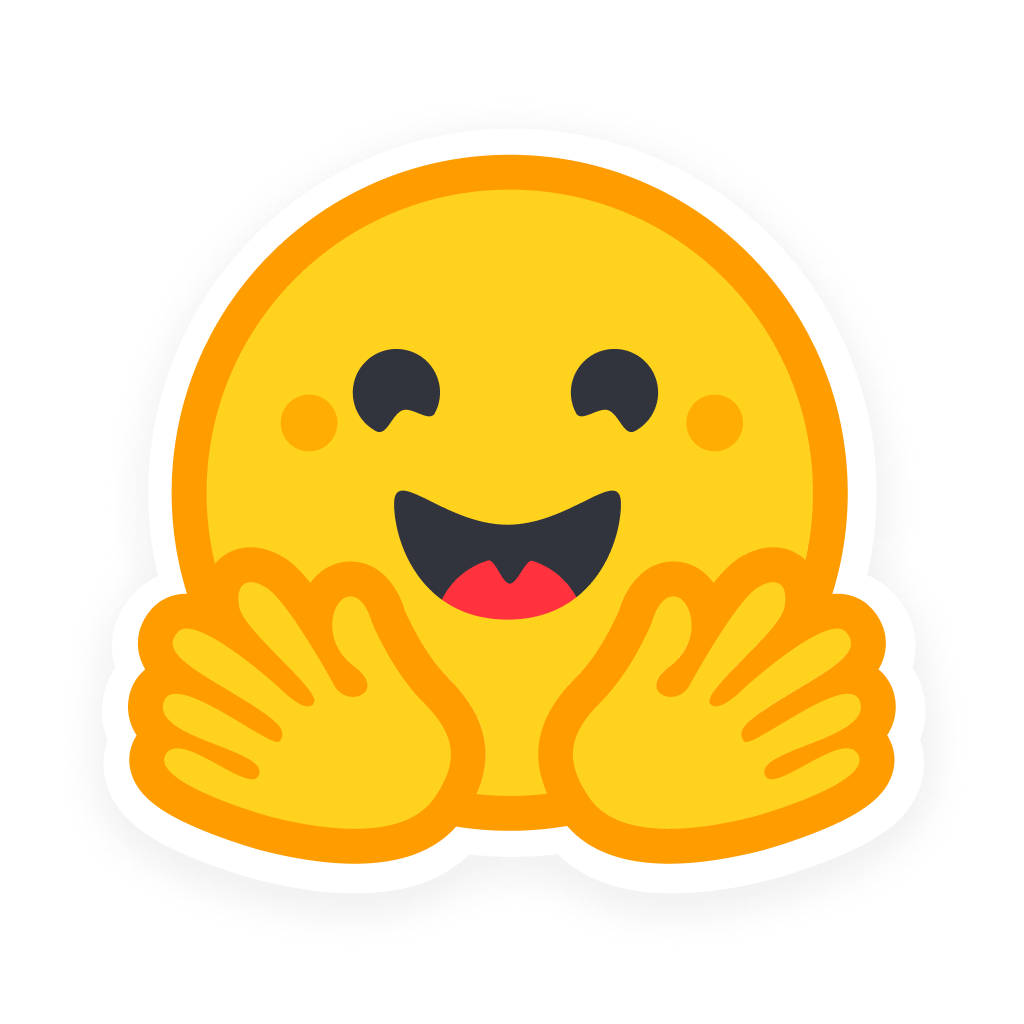}}}

\colorlet{cameranew}{blue}

\title{Semantic Motion Anchors: Bridging Motion and Meaning in Co-Speech Gestures}

\author{
\textbf{Varsha Suresh}\textsuperscript{1,2\textasteriskcentered},
\textbf{Mohammad Mahdi Abootorabi}\textsuperscript{4,5,6\textasteriskcentered},\\
\textbf{Mohamed Salman}\textsuperscript{3}, 
\textbf{M.~Hamza Mughal}\textsuperscript{1,2},\\
\textbf{Christian Theobalt}\textsuperscript{1,2,3},
\textbf{Ashwin Ram}\textsuperscript{2,3},
\textbf{Jürgen Steimle}\textsuperscript{2,3},
\textbf{Vera Demberg}\textsuperscript{1,2,3}\\ [0.5em]
\textsuperscript{\textasteriskcentered} \textbf{Equal contribution}\\ \\
\textsuperscript{1}MPI for Informatics, \textsuperscript{2}Saarland Informatics Campus,
\textsuperscript{3}Saarland University \\
\textsuperscript{4}University of British Columbia,
\textsuperscript{5}Vector Institute, 
\textsuperscript{6}Zuse School ELIZA \\[0.5em]
\small \texttt{\{vsuresh,vera\}@lst.uni-saarland.de},~\texttt{mahdi.abootorabi@ece.ubc.ca} \\
\small \texttt{mosa00006@stud.uni-saarland.de},~\texttt{\{ram,steimle\}@cs.uni-saarland.de} \\
\small \texttt{\{mmughal,theobalt\}@mpi-inf.mpg.de} \\[0.6em]
\small \faGlobe~Project page: \url{https://semantic-motion-anchors.github.io/}\\
\small \faGithub~Code: \url{https://github.com/aboots/Semantic_Motion_Anchors}\\
\small \hflogo~Data: \url{https://huggingface.co/datasets/aboots/SEMANTIX}
}

\begin{document}
\maketitle
\begin{abstract}

Learning a shared representation between spoken text and gesture is central to co-speech gesture retrieval, synthesis, and understanding, but remains challenging for semantically meaningful gestures whose communicative intent is not captured by motion alone. 
Direct contrastive alignment between transcripts and continuous motion embeddings often overemphasizes low-level kinematics and misses the symbolic content of semantic gestures. 
We propose \emph{semantic motion anchors}, natural-language abstractions of gesture motion capturing physical form and communicative intent. Our method discretizes 3D gestures into body-hand motion primitives, verbalizes them into structured descriptions, and grounds them in the transcript to provide auxiliary contrastive supervision. 
On BEAT2, our method improves text-to-gesture R@1 by 8.2\% over a direct text-motion baseline and outperforms prior retrieval approaches on text to gesture and gesture to text retrieval directions. 
Beyond aggregate retrieval metrics, semantic motion anchor supervision helps retrieve gestures that are semantically meaningful for the spoken query, rather than defaulting to generic motion patterns. A downstream retrieval-augmented gesture generation study showed that users significantly preferred gestures retrieved by our approach over a retrieval-augmented generation baseline, demonstrating that semantically grounded retrieval translates to gestures that better convey communicative intent in downstream generation. Code and data are publicly available.

 %
%
%
\end{abstract}



\section{Introduction}

Gestures are a core channel of human communication. Semantically meaningful co-speech gestures can complement or reinforce spoken content, make communication more effective, and are central to tasks such as co-speech gesture synthesis and understanding \citep{nyatsanga2023comprehensive}. 
A key requirement underlying these tasks is learning a shared space between language and gestures that meaningfully aligns raw motion sequences with spoken language.

Learning such a space that adequately captures semantic gestures, however,  remains challenging.
Approaches that directly map raw motion sequences with spoken text often fail to capture higher-level semantics, instead learning averaged representations that are dominated by frequent beat gestures \citep{nyatsanga2023comprehensive,zhi2023livelyspeaker,ao2023gesturediffuclip,liu2025semgesture,hegde2025understanding,mughal2025retrieving}. 
This is because semantic gestures are sparse and lie in the long tail of natural human motion distributions \citep{nyatsanga2023comprehensive}, making them underrepresented despite their importance for conveying communicative intent.
Consequently, retrieval-based approaches have been proposed to inject semantically relevant gestures into the generation process \citep{zhang2024semantic,mughal2025retrieving}. 
However, these retrieval strategies are typically based on heuristic or rule-based matching, or more recently, on learning raw motion-to-text mappings specifically for semantic gestures \citep{hegde2025understanding}, and thus remain limited in effectively modeling semantic gesture alignment.

A central challenge underlying these approaches lies in how gestures and spoken language are represented and mapped. Most existing methods learn motion embeddings under reconstruction-based objectives, which emphasize low-level kinematic features, but these are often not directly aligned with communicative intent, which is highly relevant for semantic gestures. 
For instance, a semantic gesture conveying enumeration (“first, second, third”) can differ significantly in articulation across speakers, yet express the same meaning.
Conversely, semantic gestures with similar motion patterns may encode entirely different intents depending on discourse context. 
This mismatch highlights a core limitation of current approaches: they conflate similarity in physical motion with similarity in semantic meaning, making it difficult to learn representations that generalize across the sparse and diverse space of semantic gestures. 

In this work, we argue that semantic gesture retrieval should not rely solely on directly mapping spoken text to continuous motion space, but should be supported by a semantically relevant abstraction that can better link the two modalities.
To this end, we introduce \textit{semantic motion anchors}: structured natural-language descriptions that re-express motion abstracted in terms of physical form and communicative intent. 
Here, physical form refers to gesture-relevant properties such as handedness, spatial position, motion trajectory, and hand configuration \citep{kipp2005gesture}, rather than raw frame-level joint coordinates. Communicative intent captures the gesture’s contextual function, such as listing, self-reference and uncertainty. Together, these anchors preserve motion aspects that matter for interpretation while reducing sensitivity to low-level kinematic variation that may be irrelevant for learning the shared space.

Our approach consists of three main components: First, we train a two-stream RVQ-VAE \citep{van2017neural,liu2024emage} to compress continuous 3D gesture sequences into discrete motion tokens, and deterministically map each primitive to a structured natural-language fragment describing observable spatial and kinematic properties such as hand position, movement direction, handedness, and hand configuration taken from \cite{kipp2005gesture}.
Second, we use an LLM to compose these token-level descriptions with the speech transcript into semantic motion anchors for each gesture. 
%
%
Third, we use these anchors as auxiliary supervision in contrastive text-gesture motion retrieval training. 
We hypothesize that this allows the model to learn relevant details  required for retrieval, i.e., gesture form and function, making the mapping between spoken language and motion more semantically grounded.

On BEAT2~\citep{liu2024emage}, our method improves text-to-gesture R@1 from 39.1 to 42.3, a +3.2 point absolute gain, corresponding to an 8.2\% relative improvement over the direct text-gesture motion baseline. In a downstream retrieval-augmented gesture generation study, users significantly preferred gestures retrieved by our approach over those from RAG-Gesture~\cite{mughal2025retrieving} (72.2\% vs. 27.8\%, $p<0.0001$), demonstrating that semantically grounded retrieval translates to gestures that better match communicative intent in practice.

Our contributions are: \textbf{(i)} We introduce \textit{semantic motion anchors} for text-gesture retrieval, representing co-speech gestures through natural-language descriptions of form and intent. \textbf{(ii)} We propose an anchor-supervised contrastive learning framework that uses motion-token verbalization and transcript grounding to improve language-gesture alignment.
\textbf{(iii)} We release \textsc{Semantix}, a dataset of 878 human-annotated TED and BEAT2 clips with gold form and intent descriptions for evaluating semantic gesture understanding.
\textbf{(iv)} We demonstrate gains on BEAT2 retrieval, TED--BEAT2 cross-dataset semantic retrieval, and downstream user preference against RAG-Gesture.

\vspace{-1em}

\section{Related Work}

\subsection{Co-Speech Gestures}

Co-speech gestures are an integral part of spoken communication and convey meaning jointly with speech \citep{mcneill1992hand,kendon2004gesture}. Gesture studies commonly distinguish representational gestures, such as iconic, metaphoric, and deictic gestures, from beat gestures, which primarily mark rhythm \citep{mcneill1992hand,mcneill2005gesture}. This distinction is important for computational modeling: beat gestures are frequent and relatively well captured by speech-synchronized motion models, whereas semantic gestures are sparse, context-dependent, and often lie in the long tail of natural gesture distributions \citep{nyatsanga2023comprehensive,ram2025gesturecoach,mughal2025retrieving}.

Recent co-speech gesture generation methods notably improve naturalness and temporal alignment by modeling speech-conditioned motion, but often produce generic or beat-dominated gestures \citep{nyatsanga2023comprehensive,zhi2023livelyspeaker,ao2023gesturediffuclip}. 
To address this, semantics-aware approaches introduce language-motion alignment, semantic planning, or retrieval-augmented generation to produce gestures that better match discourse meaning \citep{ao2023gesturediffuclip,zhi2023livelyspeaker,zhang2024semantic,mughal2025retrieving,liu2025semgesture, ram2025gesturecoach}.
These methods suggest that semantic grounding is important for communicatively meaningful generation.
However, because their primary focus is synthesis, retrieval is often treated as an intermediate step and implemented using rule-based or heuristic matching. 
In contrast, we study semantic gesture retrieval as the main task, focusing on how spoken language and gesture motion can be aligned through natural-language descriptions of gesture form and communicative intent.

\subsection{Text to Motion Retrieval}

Text-to-motion retrieval learns a shared embedding space between natural-language descriptions and motion sequences \citep{petrovich2023tmr}. Existing methods such as TMR and MotionGPT are typically developed on standard human motion benchmarks, where captions directly describe the performed action, making the language-motion relation relatively literal \citep{guo2022generating,petrovich2023tmr,jiang2023motiongpt,petrovich2022temos}.

Co-speech gesture retrieval is more implicit: the transcript rarely describes the gesture, but instead provides discourse context from which the gesture function must be inferred. Thus, direct transcript-motion alignment can conflate low-level kinematic similarity with communicative similarity. JEGAL~\cite{hegde2025understanding} is closest to our setting, as it learns gesture-language alignment for co-speech gestures through direct multimodal contrastive learning \citep{hegde2025understanding}. Unlike direct contrastive alignment, our approach introduces an intermediate linguistic abstraction: gestures are verbalized into physical-form and communicative-intent anchors. This allows retrieval to be shaped by symbolic gesture content rather than only raw motion similarity. Our focus on text-motion alignment is grounded in prior gesture research showing that semantic gestures (iconic, metaphoric, deictic) are primarily linked to linguistic content \cite{mcneill1992hand}, and a majority of co-speech generation works \cite{mughal2025retrieving,liu2025semgesture,zhang2024semantic} use text as the primary source of meaning, and we follow the same design choice.

\vspace{-0.5em}
\section{Semantic Motion Anchor Supervision for Text-to-Gesture Retrieval}

Text-to-gesture retrieval is defined over paired samples 
$(X_i, y_i)$, where $X_i \in \mathbb{R}^{T \times D}$ is 
a 3D gesture sequence ($T$ frames, $D{=}114$ per-frame pose dimensions 
over 38 upper-body joints) and $y_i$ is the spoken transcript.
Standard contrastive retrieval directly aligns $X_i$ and 
$y_i$; we additionally construct a semantic motion anchor 
$a_i$ for each 
pair, generated from motion and transcript 
and used only during training as an auxiliary contrastive 
supervision.

\subsection{Semantic Anchor Generation}
\label{semantic_anchor_generation}
The goal of semantic motion anchor generation is to convert continuous gesture motion into a compact natural-language description that captures both what the gesture looks like and what communicative role it serves in context. 
We generate each anchor in three steps: motion tokenization, token verbalization, and transcript-grounded reasoning.

\textbf{Motion Tokenization:} Given a gesture sequence $X_{i}$, we first compress the continuous 3D motion sequence into a sequence of discrete motion tokens using a two-stream RVQ-VAE ~\citep{van2017neural}. 
We use the upper body coordinates as we are interested in hand gestures and it is split into body and hand streams, $X_{i} = (X_{i}^{\text{body}}, X_{i}^{\text{hand}})$ and encoded separately and quantized using separate codebooks following \cite{liu2024emage}. 
This allows each motion sequence to be converted into a set of discrete motion tokens $q_{i}=(q_{i1},q_{i2}, \ldots ,q_{iN})$ where each token $q_{ij} = (q^{\text{body}}_{ij}, q^{\text{hand}}_{ij})$ represents an 8-frame segment 
encoded jointly across body and hand streams. Further architecture, training details and hyperparameter tuning are provided in Appendix~\ref{appx:rvq-vae}.

\textbf{Token verbalization via Gesture Attribute Extraction:} Each token $q_{ij}$ is mapped to a structured natural-language fragment $d_{ij}$ describing its observable physical properties for both hands and body.  
This step is grounded in \citet{kipp2005gesture}'s distinction between describing the visible form of a gesture and interpreting its communicative function. In our case, token verbalization performs the first step: it records what the hands and arms do, without using the transcript or inferring the gesture’s meaning. Following \citet{kipp2005gesture}'s gesture annotation dimensions, we describe each token in terms of handedness, spatial location, movement trajectory, palm orientation, and coarse hand shape.

We use these dimensions as the basis for our physical-form representation and automatically derive them from 3D skeleton geometry deterministically using numerical coordinates.  
From the body stream, we derive body-relative hand location and movement properties — wrist height, depth relative to the torso, horizontal placement, elbow bend, 
arm reach, and motion direction — capturing where the hands are placed and how they move in gesture space.
From the hand stream, we extract coarse palm orientation and hand shape, including whether the palm faces inward or outward and whether the hand is open, relaxed, curled, or pointing. 

These attributes are mapped to a chunk-level 
natural-language fragment $d_{ij}$ via a deterministic template 
function $g_{\text{temp}}$; for example, a token may be 
verbalized as ``right hand rises to chest level with an open 
palm.'' 
Concatenating these fragments in temporal order gives a 
physical motion narrative for the full gesture, 
$m = \text{concat}(d_{i1}, \dots, d_{iN})$, which preserves 
visible gesture form at a higher level of abstraction than raw 
joint coordinates. 
Full details of the geometric extraction 
rules are provided in Appendix~\ref{appx:verbalisation}.

\textbf{Transcript-Grounded Reasoning:} While the token-level motion narrative $m$ captures fine-grained physical form, it does not encode the communicative role of the gesture, which is defined relative to the spoken context.
%
%
We therefore ground $m$ in $y_i$ using GPT-5.4 via a four-stage structured 
reasoning procedure (handedness, motion, intent, 
verification), described below.

The prompt decomposes semantic motion anchor generation into four internal checks. First, \textbf{handedness}: the LLM determines whether the meaningful gesture is performed by one hand or both hands, using both text cues and asymmetries in the motion narrative. Second, \textbf{motion}: it maps the physical motion narrative into a concise spatial description, including gesture level, motion path, hand relation, palm orientation, and hand shape when supported by the motion evidence. Third, \textbf{intent}: it infers the communicative function of the gesture from the transcript, using functions such as emphasis, listing, enumeration, contrast, uncertainty, self-reference, other references, discourse, temporal reference, progression, relativity, emotion, negation, quantification or symbolic depiction. Fourth, \textbf{verification}: it checks that the inferred handedness, motion, and intent are mutually consistent before producing the final description.

The resulting semantic motion anchor $a$ is a compact natural language description that jointly encodes gesture form and function, e.g., ``Right hand rises to chest level with open palm, emphasizing the increase described in speech.'' We use the terms semantic motion anchors and semantic anchors interchangeably throughout the paper. We have provided the full prompt in Appendix~\ref{prompt:generation} and prompt sensitivity analysis in Appendix~\ref{appx:semantic-anchor-generation}.

\begin{figure}[t]
    \centering
    \includegraphics[width=0.9\textwidth]{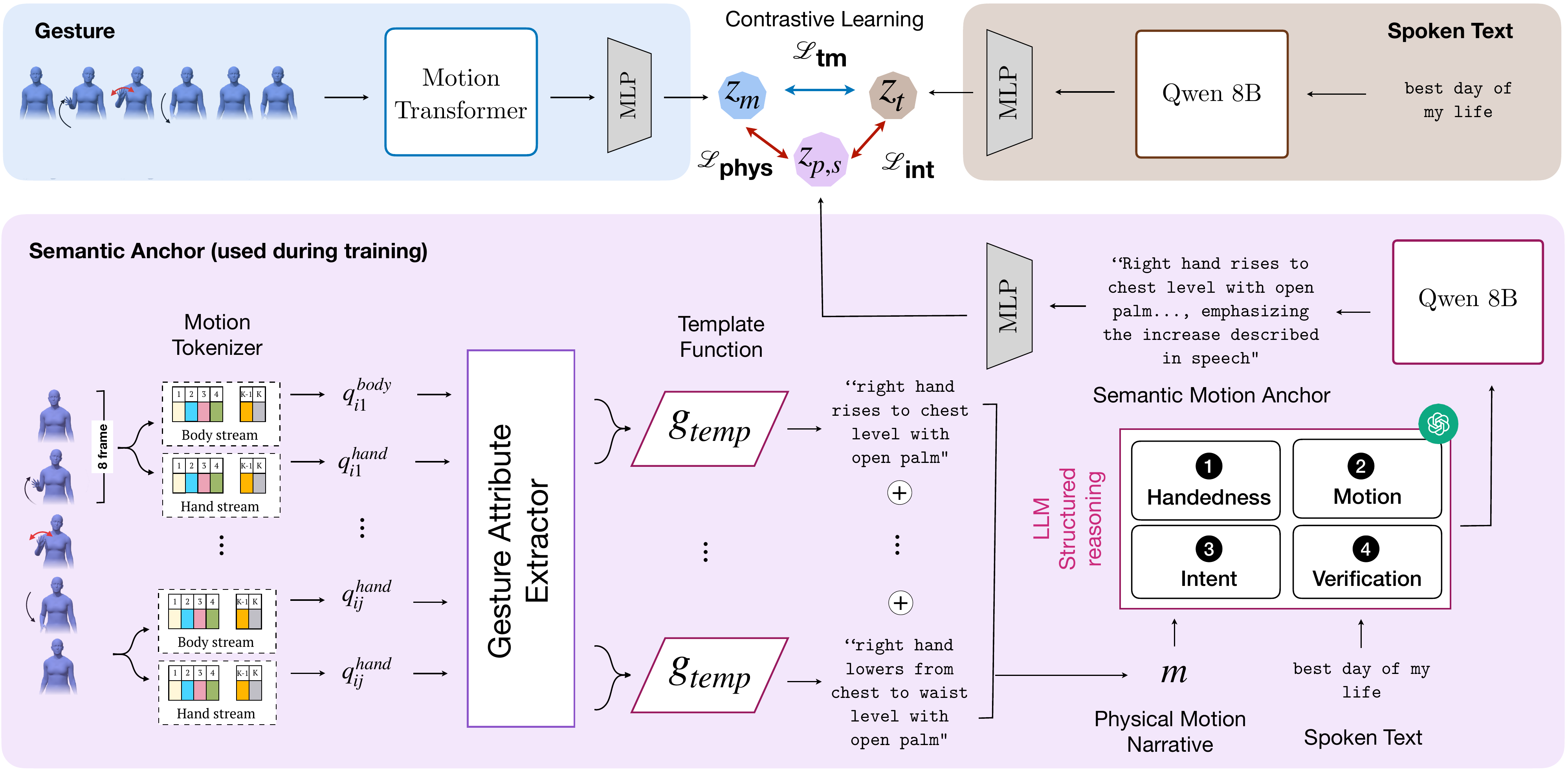}
    \caption{Overview of the proposed framework.  \textbf{Top:} The retrieval model maps transcripts and gesture motion into a shared space via contrastive learning. \textbf{Bottom:} Semantic motion anchor generation converts continuous 3D motion into discrete tokens, verbalizes them into physical-form descriptions via $g_{\text{temp}}$, and grounds them in the transcript using an LLM to produce semantic motion anchors used as auxiliary supervision during training.
    }
    \label{fig:pipeline}
    \vspace{-1em}
\end{figure}

\subsection{Anchor-supervised Contrastive Learning}
\label{sec3-2}

Figure~\ref{fig:pipeline} illustrates the full framework. Each training sample $(X_i, y_i, a_i)$ includes a semantic motion anchor decomposed into two complementary components for modality-matched supervision: $a^{phys}_i$, describing the physical form of the gesture, and $a^{int}_i$, describing its communicative intent. The decomposition is performed via a zero-shot prompt to an LLM (Qwen3-8B \cite{yang2025qwen3}); details are in Appendix~\ref{app:anchor_split}. All language inputs are embedded 
by a shared frozen text encoder $g_{\text{text}}$ 
(Qwen3-Embedding-8B; \cite{yang2025qwen3}), and the motion 
$X_i$ is encoded by a trainable transformer $f_{\text{mot}}$. 
Three learned projection MLPs map all representations into a 
shared $d$-dimensional retrieval space:

\begin{equation}
\mathbf{z}_t = \pi_{\text{tr}}\!\left(g_{\text{text}}(y_{i})\right), \quad 
\mathbf{z}_p = \pi_{\text{an}}\!\left(g_{\text{text}}(a^{\text{phys}}_{i})\right), \quad
\mathbf{z}_s = \pi_{\text{an}}\!\left(g_{\text{text}}(a^{\text{int}}_{i})\right), \quad
\mathbf{z}_m = \pi_{\text{mot}}\!\left(f_{\text{mot}}(X_{i})\right)
\end{equation}
with all outputs $\ell_2$-normalised. The transcript and anchor 
components share the frozen encoder $g_{\text{text}}$ but use 
separate projectors: $\pi_{\text{tr}}$ for the transcript and 
a single shared $\pi_{\text{an}}$ for both anchor components.

\paragraph{Training objective.}
Each $\mathcal{L}$ term denotes the symmetric InfoNCE loss with 
learnable temperature. The full objective combines four contrastive 
terms:
\begin{equation}
\label{eq:main}
\mathcal{L} = \mathcal{L}_{\text{tm}}(\mathbf{z}_t, \mathbf{z}_m) 
+ \lambda_p \, \mathcal{L}_{\text{phys}}(\mathbf{z}_p, \mathbf{z}_m) 
+ \lambda_s \, \mathcal{L}_{\text{int}}(\mathbf{z}_s, \mathbf{z}_t) 
+ \lambda_b \, \mathcal{L}_{\text{br}}(\mathbf{z}_p, \mathbf{z}_s),
\end{equation}
with auxiliary weights $\lambda_p, \lambda_s, \lambda_b$.
$\mathcal{L}_{\text{tm}}$ is the primary retrieval objective, aligning 
transcript queries to gesture motions. $\mathcal{L}_{\text{phys}}$ 
anchors the motion branch to its physical-form description, preserving 
visually grounded structure over speaker-specific variation. 
$\mathcal{L}_{\text{int}}$ anchors the transcript branch to the 
communicative-intent description, extracting gesture-relevant content 
from noisy speech context. $\mathcal{L}_{\text{br}}$ regularizes the 
shared anchor space at low weight.

\paragraph{Modality-matched supervision.}
Motion is supervised by descriptions of \emph{how the gesture looks}, while the transcript is supervised by descriptions of \emph{what the gesture means}. We route each abstraction to its corresponding modality through a shared anchor projector. Training proceeds in two stages: first, we train with only $\mathcal{L}_{\text{tm}}$ to establish the retrieval space; then, we fine-tune the projections and motion encoder with the full objective, initializing $\pi_{\text{an}}$ fresh so anchor supervision acts as structured regularization rather than replacing the main retrieval task.
\section{Evaluation of Semantic Motion Anchor Generation}
\label{sec:evaluation}

\subsection{Gold Semantic Motion Anchor Annotation}
\label{sec:quality_evaluation}

We evaluate generated anchor quality through comparison with human expert annotation and automated assessment.

\textbf{Dataset:} To support anchor quality evaluation and semantic gesture understanding, we introduce \textsc{Semantix}, a human-annotated dataset of 878 semantic gesture clips from TED Expressive and BEAT2, each paired with gold descriptions of physical form and communicative intent. To obtain these samples, we begin with the annotated gesture regions from ~\citet{ram2025gesturecoach}, which were collected on a subset of 10 videos from TED Expressive \cite{liu2022learning}. Each region includes the corresponding transcript segment and associated video clip of the semantic gesture. We annotate gold semantic gesture descriptions for 778 such regions from the training set following the gesture annotation schema of ~\citet{kipp2005gesture}. Each description specifies the gesture’s handedness, hand shape, hand orientation, spatial position, and motion trajectory, and concludes with a short phrase describing its communicative intent. To establish annotation guidelines, a primary annotator first labeled 231 samples. A second expert independently reviewed these descriptions and either accepted or revised each annotation. The reviewed annotations had a mean word-level Levenshtein distance of 0.72 with the originals. The remaining TED samples were then annotated by the primary annotator using the finalized guidelines. We further annotated 100 semantic gesture samples from the BEAT2 dataset using the same schema. Gold annotations are never used to train the retrieval model, which is trained on automatically generated (silver) anchors for all 15,395 BEAT2 training samples.


\textbf{LLM-as-judge validation:} To enable scalable quality assessment, we validated automated LLM-based evaluation with GPT-5.4, using the prompt in Appendix \ref{prompt:evaluation}. The model compares each generated description against a gold reference and assigns separate 1--5 Likert scores for physical gesture similarity and communicative intent similarity.

\textbf{Semantic motion anchor quality:} We evaluate agreement between LLM-based automatic evaluation and human judgments using Spearman rank correlation. An expert annotator rated 100 generated anchors against gold annotations, using a random subset of 50 TED and 50 BEAT2 examples. Each anchor was evaluated on two 5-point Likert scales: Pose Score, measuring the accuracy of the physical gesture description from 1 = incorrect to 5 = perfect, and Intent Score, measuring the accuracy of the communicative function from 1 = wrong to 5 = correct. For TED, we observe strong correlations between LLM and human judgments for both pose ($\rho$ = 0.887, p < 0.001) and intent ($\rho$ = 0.810, p < 0.001). The LLM scores are slightly lower than the human scores on average, with mean pose scores of 3.44 for the LLM and 3.75 for the human annotator, and mean intent scores of 4.20 for the LLM and 4.48 for the human annotator. For BEAT2, we observe similarly strong correlations for pose ($\rho$ = 0.942, p < 0.001) and intent ($\rho$ = 0.947, p < 0.001). Again, mean scores assigned by LLMs are slightly lower (3.34 for pose, 4.42 for intent) than scores assigned by humans (3.77 for pose and 4.60 for intent). These results suggest that the LLM-as-judge captures relative ranking trends consistent with human evaluation on this validation subset.
\section{Text-Gesture Retrieval}

\subsection{Retrieval Setup}

We train the retrieval model on BEAT2 and evaluate it on the BEAT2 test split, with TED used only for out-of-domain evaluation. Each sample contains a transcript window, the corresponding 3D upper-body motion sequence, and the generated physical-form and communicative-intent anchors. BEAT2 is split into 90\% training, 5\% validation, and 5\% test sets ($N_{\text{train}}{=}15{,}395$, $N_{\text{val}}{=}855$, $N_{\text{test}}{=}856$). Model selection is based on transcript-motion MRR on the BEAT2 validation split. At inference time, for text-gesture retrieval the anchor branches are discarded, and retrieval is performed only between transcript and motion embeddings using cosine similarity over the full test gallery ($N{=}856$ candidates for BEAT2; $N{=}778$ for TED). We report Recall@1, Recall@5, Recall@10, and MRR for both text-to-gesture and gesture-to-text retrieval. We compare against GestureDiffuCLIP, TMR, JEGAL, and a direct text-motion contrastive (Text Contrastive in the table) baseline under the same data splits and evaluation protocol. Full implementation details are provided in the Appendix~\ref{appx:cross-modal-implementation-details}.

\begin{table*}[b]
\centering
\resizebox{\linewidth}{!}{%
\begin{tabular}{lcccccccc}
\toprule
 & \multicolumn{4}{c}{\textbf{Gesture $\rightarrow$ Text}} & \multicolumn{4}{c}{\textbf{Text $\rightarrow$ Gesture}} \\
\cmidrule(lr){2-5} \cmidrule(lr){6-9}
\textbf{Method} & R@1 & R@5 & R@10 & MRR & R@1 & R@5 & R@10 & MRR \\
\midrule
GestureDiffuCLIP~\citep{ao2023gesturediffuclip} & 32.3 & 57.4 & 66.6 & 44.0 & 33.8 & 57.5 & 67.2 & 45.1 \\
TMR~\citep{petrovich2023tmr}                    & 37.4 & 57.7 & 65.7 & 47.0 & 39.1 & 58.7 & 66.4 & 48.6 \\
JEGAL~\citep{hegde2025understanding}            & 36.6 & 58.4 & 66.6 & 47.0 & 39.3 & 59.3 & 66.8 & 48.9 \\
\midrule
Text Contrastive                               & 37.2 & 57.5 & 65.4 & 47.0 & 39.1 & 58.7 & 66.3 & 48.5 \\
\textbf{Text Contrastive w Semantic Anchors (Ours)} & \textbf{41.8}$^{\dagger}$ & \textbf{62.0}$^{\dagger}$ & \textbf{68.9}$^{\dagger}$ & \textbf{51.4}$^{\dagger}$ & \textbf{42.3}$^{\dagger}$ & \textbf{62.5}$^{\dagger}$ & \textbf{69.5}$^{**}$ & \textbf{51.9}$^{\dagger}$ \\
\bottomrule
\multicolumn{9}{l}{\footnotesize $^{*}p<0.05$, $^{**}p<0.01$, $^{\dagger}p<0.001$ compared to Text Contrastive using paired t-test}
\end{tabular}}
\caption{%
  Bidirectional text–gesture retrieval evaluated over the full test gallery of BEAT2 N=856 candidates. Best results are \textbf{bold}.
}
\label{tab:main_results}
\end{table*}
\vspace{-0.5em}
\subsection{Main Retrieval Results}

\paragraph{Does semantic motion anchor supervision improve retrieval?}

Table~\ref{tab:main_results} compares our method against existing approaches on BEAT2 across 7 random seed runs. Our method outperforms all baselines on both retrieval directions. Relative to the strongest prior baseline JEGAL, gesture-to-text retrieval improves by 14.2\% in R@1 and 9.4\% in MRR, while text-to-gesture retrieval improves by 7.6\% in R@1 and 6.1\% in MRR. Gains are consistent but smaller at higher recall cutoffs, with R@5 and R@10 improving by 3.5–6.2\% across both directions, reflecting that the primary benefit of semantic motion anchor supervision is concentrated at the top rank, precisely where co-speech gesture systems must commit to a single retrieval decision. Beyond aggregate metrics, the cumulative rank distribution (in Appendix~\ref{appx:cumrank}) confirms that this advantage is concentrated in the low-ranks.

\begin{figure}[h]
\centering
\begin{minipage}[c]{0.52\linewidth}
    \centering
    \small
    \resizebox{\linewidth}{!}{
\resizebox{\linewidth}{!}{%
\begin{tabular}{lcccc}
\toprule
 & \multicolumn{2}{c}{\textbf{Gesture $\rightarrow$ Text}} 
 & \multicolumn{2}{c}{\textbf{Text $\rightarrow$ Gesture}} \\
\cmidrule(lr){2-3} \cmidrule(lr){4-5}
\textbf{Method} & R@5 & MRR & R@5 & MRR \\
\midrule
No Anchor 
& 57.54 & 46.98 
& 58.74 & 48.49 \\

Random Anchor
& 60.35 & 50.02
& 60.96 & 51.22 \\
Mismatched Anchors & 60.33& 50.11& 61.28 & 51.15\\
Physical Only & 60.21 & 50.01 & 60.98 & 50.92 \\
Intent Only & 62.00 & $\mathbf{51.44}$ & $\mathbf{62.53}$ & $\mathbf{52.04}$ \\

\midrule
\textbf{Semantic Anchors}
& $\mathbf{62.05}^{\dagger}$
& $\mathbf{51.44}^{\dagger}$
& $\mathbf{62.53}^{\dagger}$
& $51.85^{\dagger}$ \\
\bottomrule
\end{tabular}}

}
    \captionof{table}{Anchor ablation on BEAT2. $\dagger$ indicates significant improvement of Semantic Anchors over both Random and Mismatched Anchors using paired t-test; $p < 0.05$.}
    \label{tab:anchor}
\end{minipage}
\hfill
\begin{minipage}[c]{0.44\linewidth}
    \centering
    \includegraphics[width=\linewidth]{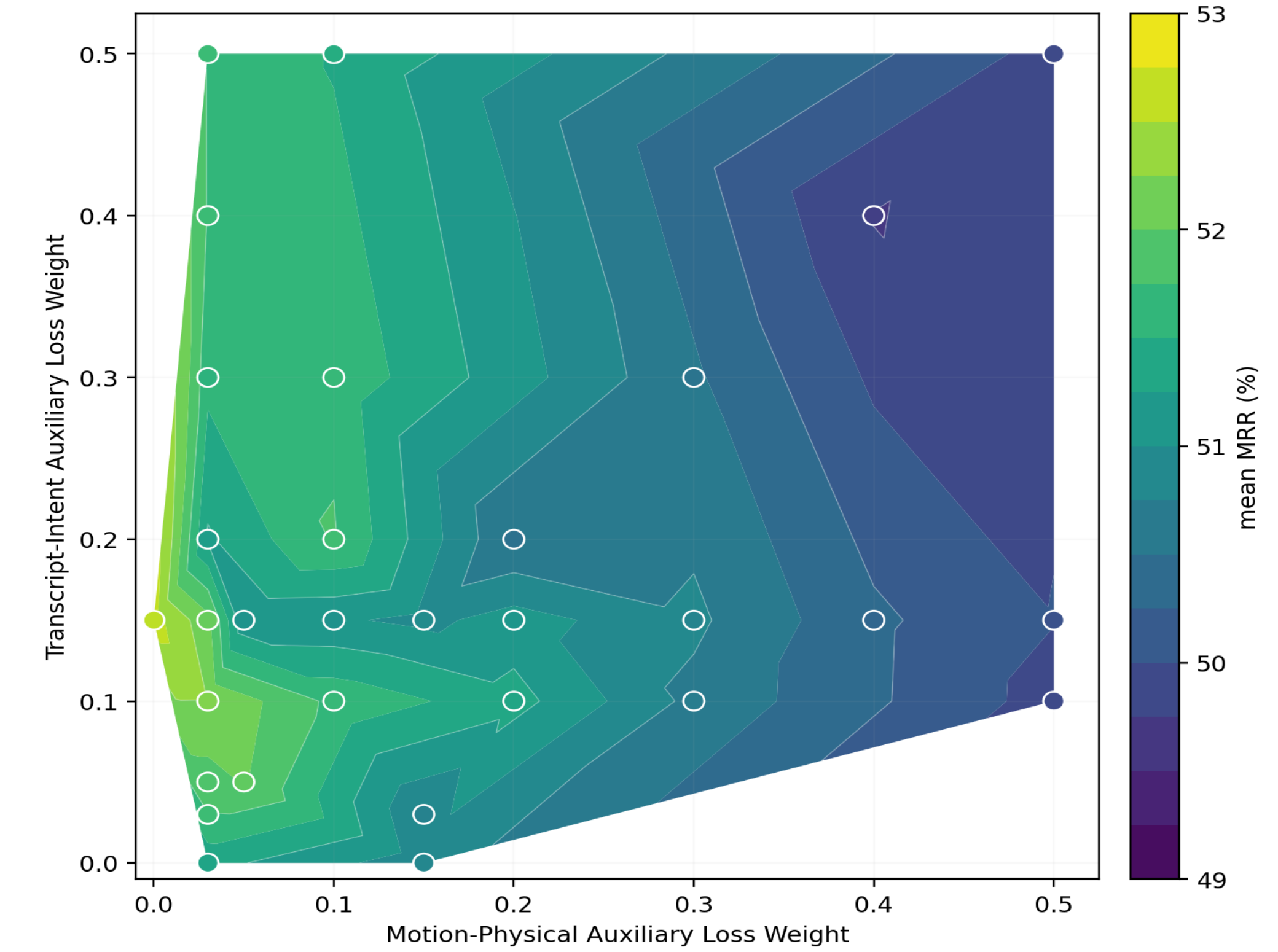}
    \captionof{figure}{Joint sensitivity of $\lambda_p$ (physical-form) and $\lambda_s$ (intent) on mean MRR (\%)}
    \label{fig:weight_sensitivity}
\end{minipage}
\end{figure}

\paragraph{Effect of anchor content on retrieval performance:}
Replacing the anchor text embeddings $g_{\text{text}}(a^{phys})$ and $g_{\text{text}}(a^{int})$ with fixed per-sample Gaussian unit vectors partially recovers the gain over the no-anchor baseline, showing that the auxiliary contrastive structure itself provides an in-domain regularization benefit. However, semantic motion anchors further improve over random targets across both R@5 and MRR ($p < 0.05$), indicating that meaningful anchor content adds supervision beyond regularization alone.
As shown in Table~\ref{tab:anchor}, mismatched anchors perform similarly to random anchors, while correctly paired semantic anchors improve R@5 and MRR in both retrieval directions ($p < 0.05$). This suggests that the gains beyond regularization depend on matching meaningful anchor content to the correct samples.
Single-branch ablations also show that the in-domain gain comes mainly from intent supervision, while the physical branch aligns physical-form descriptions with motion, which lets them serve as cross-dataset motion proxies (Appendix~\ref{appx:cross8}).
The intent anchor is not a paraphrase of the transcript: it is generated jointly from the transcript and the motion narrative with a motion-consistency check (Sec.~\ref{semantic_anchor_generation}), and mismatched anchors give no gain over random ones, so the benefit of $\mathcal{L}_{\text{int}}$ requires each anchor to be paired with its own sample.
Details and the full table are provided in Appendix~\ref{app:random_anchor} and Appendix~\ref{appx:full-table-anchor-content}.

\paragraph{Does physical-form or intent supervision matter more?}
Figure~\ref{fig:weight_sensitivity} shows the joint sensitivity of $\lambda_p$ (physical-form) and $\lambda_s$ (intent) on mean MRR.
Performance degrades consistently as $\lambda_p$ increases, regardless of $\lambda_s$, while the intent branch remains stable across a broad range of $\lambda_s$ values. This suggests the motion branch is more susceptible to over-regularization than the intent branch. Peak MRR is achieved at small $\lambda_p$ ($0.01$--$0.05$) with moderate $\lambda_s$ ($0.10$--$0.15$). Marginal sensitivity curves with one weight fixed are provided in Appendix~\ref{app:marginal_sensitivity}.

\subsection{Qualitative analysis: Analyzing Semantic Alignment of Retrieved Gestures}
\label{sec:qualitative}
\vspace{-0.3em}
\paragraph{Semantic label match rate:} Standard Recall@K treats only the paired ground-truth gesture as correct. However, co-speech gestures are many-to-many: a different gesture can still express the same communicative intent. We do this analysis on the text to gesture retrieval on the test gallery of BEAT2.

\begin{wraptable}{r}{0.46\linewidth}
\centering
\small
\resizebox{\linewidth}{!}{
\begin{tabular}{lrrrr}
\toprule
Category & $n$ & Ours & Text & Random \\
\midrule
Emphasis           & 250 & \textbf{57.2} & 52.8 & \textbf{57.2} \\
Discourse          & 227 & \textbf{62.1} & 57.7 & 61.7 \\
Other-reference    & 111 & \textbf{59.5} & 56.8 & 58.6 \\
Self-reference     &  92 & \textbf{52.2} & \textbf{52.2} & 44.6 \\
Symbolise          &  68 & \textbf{48.5} & 44.1 & 44.1 \\
Contrast           &  24 & \textbf{50.0} & 41.7 & \textbf{50.0} \\
Emotion            &  16 & \textbf{56.2} & 43.8 & 50.0 \\
Uncertainty        &  15 & \textbf{66.7} & 53.3 & 60.0 \\
Temporal reference &  12 & \textbf{50.0} & 33.3 & \textbf{50.0} \\
Quantification     &  11 & \textbf{45.5} & 27.3 & 27.3 \\
Negation           &  10 & 50.0 & \textbf{60.0} & 50.0 \\
Relativity         &   9 & 66.7 & 55.6 & \textbf{77.8} \\
Enumeration        &   6 & 33.3 & 33.3 & 33.3 \\
Progression        &   3 & 0.0 & 0.0 & 0.0 \\
Listing            &   2 & 50.0 & 50.0 & 50.0 \\
\midrule
\textbf{Overall}   & 856 & \textbf{56.9} & 52.6 & 55.1 \\
\bottomrule
\end{tabular}}
\captionof{table}{Semantic label match rate (\%) 
on BEAT2 test set.}
\label{tab:label_match}
\vspace{-0.6em}
\end{wraptable}
We compute semantic label match rate, which measures whether the top-1 retrieved gesture shares the same intent label as the ground-truth gesture. Unlike Recall@1, this metric rewards \emph{semantic alignment}: retrieving
a different gesture instance that expresses the same communicative intent
is counted as correct. The semantic intent labels for the BEAT2 test samples were extracted from the communicative intent descriptions using Qwen3-8B, with the SEMANTIX intent categories (Section~\ref{semantic_anchor_generation}).

From Table~\ref{tab:label_match} we see that the largest numerical gains appear in Quantification, Temporal reference, Uncertainty, and Emotion,  which are categories with distinctive gestural form where intent-conditioning provides the clearest signal and rarer in the data.

\begin{table*}[t]
\centering
\small
\begin{tabular}{p{0.46\linewidth} | p{0.46\linewidth}}
\toprule
\textbf{Example 1 — \textit{Emotion}} & \textbf{Example 2 — \textit{Self-reference}} \\
\midrule
``\ldots I miss my relatives and friends the most, \textbf{I missed the feelings} that I'm at home \ldots'' 
& 
``\ldots when \textbf{I\textquotesingle m free} I like to listen to music or watch documentary movies\ldots'' \\
\midrule
\textbf{GT:} Both hands rise from the waist to chest level and spread outward, as if opening up a space. Conveys a broad, heartfelt feeling of home and familiarity.
&
\textbf{GT:} Both hands make a small outward spread low near the waist, as if opening up a space, to present what the speaker does in their free time. \\
\midrule
\textbf{Ours:} Both hands rise from low to chest level and hover with open, relaxed palms, drawing an inward personal feeling into focus. Conveys longing and attachment to the feeling of being at home.
&
\textbf{Ours:} One hand lifts from waist to chest level and opens slightly toward the speaker, while the other remains low and still, to refer to spending time with oneself in a personal, inviting way. \\
\midrule
\textbf{Text Contrastive:} One hand moves outward and downward from shoulder to chest level, as if briefly presenting something to the side. Illustrates the idea of pausing to notice sensory details.
&
\multirow{2}{=}{\textbf{Text Contrastive \& Random:} One hand makes a small downward-then-upward beat near waist level at the side, marking the rhythm of the phrase to emphasise a habitual action in the speaker's routine.} \\
\textbf{Random:} Both hands rise to chest level, spread outward and lower with open palms, as if laying out a sequence of activities.
& \\
\bottomrule
\end{tabular}
\caption{Qualitative retrieval examples.}
\label{tab:qual_eg_main}
\vspace{-1.5em}
\end{table*}

A macro average across all 15 categories (equal weight per category) increases our margin from +4.3 points overall to +5.8 points (49.9 vs. 44.1 for Text Contrastive), confirming that aggregate improvement is not driven by the dominant Emphasis and Discourse categories. Furthermore, Table~\ref{tab:qual_eg_main} shows two examples where our model retrieves a gesture with a similar semantic intent. 
In both cases, our model retrieves a gesture that better matches the intended semantic label, while the text-only and random-anchor baselines retrieve gestures with mismatched communicative functions.
For more examples, refer to Table~\ref{tab:qual_eg_appendix} in Appendix. %

\vspace*{-0.6em}
\subsection{Cross-Dataset Generalization on TED}
\label{sec:cross_dataset}
\vspace{-0.3em}

We evaluate whether a retrieval model trained only on BEAT2 transfers to the unseen TED dataset~\cite{ram2025gesturecoach} ($N{=}778$). We focus on text-to-gesture retrieval because this is the most common retrieval direction in downstream settings such as retrieval-augmented co-speech gesture generation. In all experiments, the query is a TED transcript segment. We use two controlled gallery settings. In the first setting, \textbf{TED-to-TED}, the gallery also consists of TED gestures. This tests whether the BEAT2-trained model can be applied to a new dataset at inference time. Since both the query and gallery come from TED, we represent TED gallery gestures using only the physical-form anchor $a^{\mathrm{phys}}$, rather than transcript-derived semantic anchors, to avoid leakage from the paired transcript. In the second setting, \textbf{TED-to-BEAT2}, the query comes from TED and the gallery is the BEAT2 test set. This is a stronger out-of-domain retrieval setting: the transcript query and gesture gallery come from different datasets, while the retrieval model itself is trained only on BEAT2.
TED-to-TED is a diagnostic setting, whereas TED-to-BEAT2 reflects the practical scenario of matching queries from a new domain against an existing curated gesture gallery.

For each gallery setting, we compare two gallery representations: raw motion embeddings and semantic anchor proxies. The motion-embedding gallery tests whether the learned BEAT2 motion space transfers directly. The anchor-proxy gallery tests whether abstracting gestures into relevant physical or communicative properties provides a more transferable retrieval interface. For TED-to-TED, exact transcript–gesture pairs are available, so we report R@5 and MRR. For TED-to-BEAT2, exact pairs are unavailable, we evaluate retrieval using shared semantic-label metrics (Acc@1, Hit@5, Hit@10, MRR, label nDCG@10) and semantic-context similarity with frozen Qwen3 embeddings (BestCos@5, MeanCos@10, semantic nDCG@10). Metric details are provided in Appendix~\ref{app:cross_dataset_proxy_metrics}. Recomputing the semantic-context metrics with an independent embedding model (BGE-M3) ranked our method first on every metric in both gallery settings, which indicates that these results are not an artifact of evaluating with the same embedding family as our text encoder (Appendix~\ref{appx:bge_m3}).


\begin{minipage}[h]{0.52\linewidth}


\paragraph{TED-to-TED:} This setting isolates how the BEAT2-trained motion encoder handles TED motion. Applied directly, the encoder yields near-chance performance (Table~\ref{tab:ted_generalisation_proxies}). Although both datasets are expressed in SMPL-X, studio motion capture (BEAT2) and monocular video pose estimation (TED) produce very different motion distributions: a probe can separate the two datasets near perfectly (more details in Appendix~\ref{appx:domain_shift}). TED-to-TED therefore tests whether the learned semantic space remains usable when the raw motion encoder does not transfer.

\end{minipage}
\hfill
\begin{minipage}[h]{0.46\linewidth}
\centering
\small
\resizebox{\linewidth}{!}{%
\begin{tabular}{llcccc}
\toprule
 & & \multicolumn{2}{c}{\textbf{Gesture $\rightarrow$ Text}} 
 & \multicolumn{2}{c}{\textbf{Text $\rightarrow$ Gesture}} \\
\cmidrule(lr){3-4} \cmidrule(lr){5-6}
\textbf{Method} & \textbf{Gesture Rep.} 
& R@5 & MRR & R@5 & MRR \\
\midrule

Baseline\textsuperscript{*} & Motion embed. & 1.3 & 1.17 & 0.5 & 0.96 \\

\midrule

Text Contrastive $\dagger$ 
& $a^{phys}$
& 1.8 & 1.91 
& 2.8 & 2.64 \\

Random Anchor 
&  $a^{phys}$
& 0.5 & 0.80 
& 0.5 & 0.79 \\

Semantic Anchor
& $a^{phys}$ 
& \textbf{4.6} & \textbf{3.48} 
& \textbf{4.5} & \textbf{3.42} \\
\bottomrule
\end{tabular}

}
\captionof{table}{%
  \textbf{TED-to-TED} Diagnostic cross-dataset evaluation. {*}since motion-embed. variants perform near chance, we report only the best one. $\dagger$ indicates that motion descriptions are passed through the transcript encoder.%
}
\label{tab:ted_generalisation_proxies}
\end{minipage}
\begin{table*}[t]
\centering
\resizebox{\textwidth}{!}{%
\begin{tabular}{l ccccc cc ccc}
\toprule
& \multicolumn{5}{c}{\textbf{Semantic Label (\%)}} & \multicolumn{2}{c}{\textbf{Pairwise Win Rate of Proposed (\%)}} & \multicolumn{3}{c}{\textbf{Semantic Context (\%)}} \\
\cmidrule(lr){2-6} \cmidrule(lr){7-8} \cmidrule(lr){9-11}
\textbf{Method}
& Acc@1 & Hit@5 & Hit@10 & MRR & nDCG@10
& nDCG@10 & MeanCos@10 
& BestCos@5 & MeanCos@10 & nDCG@10 \\
\midrule
\multicolumn{11}{l}{\textbf{Setup: TED Transcripts as Query \& BEAT2 Motion Embeddings as Gallery}} \\
\midrule
Text Contrastive 
& \textbf{12.7} & \textbf{38.6} & 48.2 & \textbf{24.3} & 11.7 
& 57.2 & 54.4 
& 55.6 & 48.0 & 69.4 \\
Random Anchor 
& 12.5 & 37.1 & 49.1 & 23.9 & 12.3 
& 57.5 & 55.5 
& 55.7 & 48.1 & 69.5 \\
Mismatched Anchors
& 12.0 & 38.2 & 48.8 & 23.7 & 12.3 
& 56.8 & 55.8 
& 55.8 & 48.1 & 69.5 \\
Proposed
& 11.3 & 37.9 & \textbf{49.7} & 23.1 & 12.3 
& -- & -- 
& \textbf{55.9} & \textbf{48.3} & \textbf{69.8} \\
\midrule
\multicolumn{11}{l}{\textbf{Setup: TED Transcripts as Query \& BEAT2 Semantic Anchor Proxies as Gallery}} \\
\midrule
Text Contrastive 
& 15.4 & 37.1 & 49.2 & 26.3 & 14.0 
& 54.9 & 57.7 
& 56.9 & 48.8 & 70.5 \\
Random Anchor 
& 9.4 & 32.0 & 43.6 & 20.2 & 9.7 
& 71.3 & 71.6 
& 53.6 & 46.5 & 67.1 \\
Mismatched Anchors
& 13.2 & 33.9 & 45.1 & 23.0 & 11.4 
& 76.5 & 77.2 
& 53.7 & 46.3 & 66.6 \\
Proposed
& \textbf{17.1} & \textbf{40.6} & \textbf{52.3} & \textbf{28.0} & \textbf{14.4} 
& -- & -- 
& \textbf{57.2} & \textbf{49.4} & \textbf{71.2} \\
\bottomrule
\end{tabular}}
\caption{
Cross-dataset retrieval on \textbf{TED-to-BEAT2} setting. We compare direct BEAT2 motion embeddings with motion anchor proxies. Semantic label metrics use the semantic intent categories; semantic-context metrics use embedding similarity between retrieved gesture descriptions. Win-rate reports the fraction of queries where our method outperforms each baseline.
}
\label{tab:ted_to_beat_emb_vs_proxy}
\vspace{-1.5em}
\end{table*}

To investigate this, we use a proxy-based representation of the TED gesture gallery. To prevent transcript leakage, we represent gallery gestures using only physical-form text proxies ($a^{\mathrm{phys}}$). 
Bypassing the out-of-domain raw motion encoder with these $a^{\mathrm{phys}}$ proxies helps partially recover the retrieval performance over the text-contrastive baseline (Table~\ref{tab:ted_generalisation_proxies}). Furthermore, the random-anchor control collapses to near-chance, confirming that this cross-dataset transferability stems from capturing meaningful gesture structure rather than just in-domain regularization. Full table is provided in Appendix~\ref{appx:full-table-ted-transfer}.

\vspace*{-0.5em}
\paragraph{TED-to-BEAT2:} The TED-to-BEAT2 setting provides a stronger test of cross-dataset semantic transfer, where TED transcript queries are retrieved from the BEAT2 test gallery. Here, replacing the BEAT2 motion gallery with gesture-anchor proxies improves over the text-contrastive baseline. It also clearly outperforms the Random Anchor baseline, showing that the gain comes from the meaningful content of the anchors rather than from using any proxy representation. Mismatched Anchors also falls below the text-contrastive baseline, and our method outperforms it on 76.5\% of queries by semantic nDCG@10, which indicates that the transfer comes from correctly paired anchors rather than from the auxiliary regularizer structure alone. Results for both query--gallery directions between TED and BEAT2 are provided in Appendix~\ref{appx:cross8}.

Overall, this suggests that abstracting gestures into relevant properties makes the representation more transferable across datasets and a useful way to represent semantic gestures.

%

\vspace*{-0.4em}
\subsection{Downstream Application: Co-speech Gesture Generation} 

\begin{wrapfigure}{r}{0.42\linewidth}
    \centering
    \vspace{-1em}
    \includegraphics[width=\linewidth]{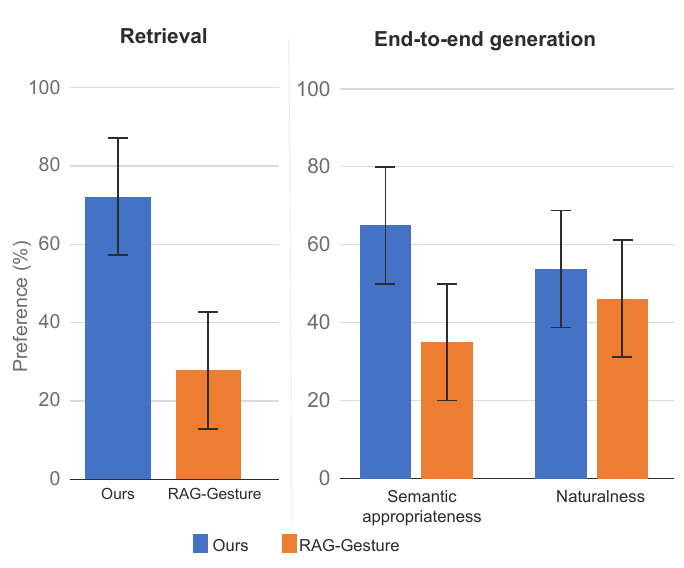}
    \caption{User study results.}
    \label{fig:user-study}
    \vspace{-0.5em}
\end{wrapfigure}

In this experiment, we evaluate retrieval quality in the setting of retrieval-augmented co-speech gesture generation. Existing RAG-based gesture generation methods, such as RAG-Gesture~\citep{mughal2025retrieving}, retrieve gesture examples through rule-based retrieval around speech/text query region and use them as guidance during generation. To demonstrate a valid application of our approach in gesture synthesis, we perform two evaluations: a perceptual user study on retrieved gestures where we replace baseline's motion retrieval step with our learned anchor-based retrieval approach and compare the motion retrievals, and an end-to-end evaluation in which retrieved gestures from both methods are passed through an identical synthesis pipeline and the resulting generations are assessed directly.

\vspace{8pt}
We evaluate the complete synthesis pipeline following RAG-Gesture’s protocol, where retrieved gestures from both methods are passed through an identical generator and assessed via two complementary sets of metrics. From the retrieval user study ($N{=}32$), we found that users preferred gestures retrieved by our approach 72.2\% of the time (vs. 27.8\% for RAG-Gesture; $W{=}11.5$, $p<0.0001$, Wilcoxon signed-rank test). From the end-to-end generation user study (16 participants, 10 gesture pairs), we found that our method was preferred for semantic appropriateness 65.0\% of the time (p = 0.0027) and for naturalness 53.8\% of the time (p = 0.295), confirming that semantically grounded retrieval improves communicative quality while maintaining comparable naturalness as shown in Figure~\ref{fig:user-study}.

\begin{minipage}[h]{0.52\linewidth}
Automatic metrics, which primarily capture naturalness and low-level motion quality, remain comparable across methods as both use the same generator. BeatAlign improves slightly over RAG-Gesture relative to the GT reference as seen in Table~\ref{tab:generation-metrics}. Details regarding the setup and questions asked during the user study are in Appendix~\ref{appx:userstudy}.
\end{minipage}
\hfill
\begin{minipage}[h]{0.46\linewidth}
\centering
\small
\resizebox{\linewidth}{!}{%
\begin{tabular}{lccc}
    \hline
    Metric & RAG-Gesture & Ours & GT \\
    \hline
    FID $\downarrow$ & 0.57 & 0.61 & {---} \\
    BeatAlign $\uparrow$ & 0.485 & \textbf{0.491} & 0.490 \\
    L1 div ($\to$GT) & 9.43 & \textbf{9.41} & 7.65 \\
    Diversity ($\to$GT) & 0.11 & 0.11 & 0.11 \\
    \hline
    \end{tabular}}
\captionof{table}{%
  Automatic gesture generation metrics.%
}
\label{tab:generation-metrics}
\end{minipage}






\vspace*{-0.7em}
\section{Conclusion}
\vspace{-0.5em}

We introduced semantic motion anchors for text-gesture retrieval: natural-language abstractions that represent gesture motion through physical form and communicative intent. Our framework converts 3D motion into upper-body tokens, verbalizes them into structured physical descriptions, and grounds them in the transcript to generate semantic motion anchors used as auxiliary supervision. Experiments on BEAT2 show that semantic motion anchor supervision improves retrieval over direct text-motion alignment. Random-anchor controls show that auxiliary supervision contributes to in-domain gains, while their collapse in cross-dataset retrieval shows the importance of meaningful semantic anchor content. Applied to retrieval-augmented co-speech gesture generation, users significantly preferred gestures retrieved by our approach over those retrieved by RAG-Gesture, which suggests that using semantic motion anchors produces better gestures that match communicative intent.

\vspace*{-0.7em}
\section{Limitations}
\label{sec:limitation}
\vspace{-0.5em}
Our semantic motion anchors capture only a subset of gesture-relevant attributes; fine-grained properties such as gesture phases and subtle finger articulation are not fully modeled. The pipeline currently consists of a simple contrastive setup; future work could explore other ways to incorporate anchors. Incorporating speech audio (e.g., prosody) is left for future work. Anchor generation introduces computational overhead and the use of a closed-source LLM. However, this is a one-time cost and is used only offline to create training anchors. As the method is trained primarily on BEAT2 and TED, it may not generalize equally across cultures, languages, or demographic groups, since gesture conventions vary significantly across these dimensions. \\

\begin{ack}
Mohammad Mahdi Abootorabi is supported by the Konrad Zuse School of Excellence in Learning and Intelligent Systems (ELIZA) through the DAAD programme Konrad Zuse Schools of Excellence in Artificial Intelligence, sponsored by the Federal Ministry of Education and Research. This work is funded in part by the Deutsche Forschungsgemeinschaft (DFG, German Research Foundation) -- GRK 2853/1 “Neuroexplicit Models of Language, Vision, and Action” -- project number 471607914, and in part by the Deutsche Forschungsgemeinschaft DFG under Grant No. 521586817 within the Priority Program SPP 2199 Scalable Interaction Paradigms for Pervasive Computing Environments.
The authors declare no competing interests.
\end{ack}

\bibliographystyle{plainnat}
\bibliography{references}

\newpage
\appendix

\section{Technical Appendices and Supplementary Material}

\subsection{RVQ-VAE Architecture and Training}
\label{appx:rvq-vae}


\textbf{Training data:} The RVQ-VAE is trained on a combined corpus of TED Expressive \citep{liu2022learning} and BEAT2 \citep{liu2024emage}, covering over 120 hours of co-speech motion tracking data. We select 38-joint upper-body skeletons in 3D (114 dimensions per frame), with TED Expressive joint coordinates estimated via ExPose \citep{choutas2020expose} and BEAT2 coordinates extracted from SMPL-X parameters \citep{pavlakos2019smplx}. We preprocess each skeleton by centering at the neck joint, scaling to unit sphere, and rotation-aligning to the torso plane via a right-up-forward coordinate frame. Finger joints are additionally translated and scaled relative to their respective wrist joints. 

\textbf{Architecture:} The model separately encodes body motion (8 joints: neck, shoulders, elbows, wrists, head) and hand articulation (30 finger joints) through independent 1D convolutional encoders with temporal downsampling factor of 8. Each stream uses three-stage residual quantization with codebook sizes (128, 128, 128) for body and (128, 64, 32) for hands. The quantized representations are decoded by a shared transposed convolutional decoder that reconstructs the full skeleton sequence.

Our two-stream RVQ-VAE takes as input an 8-frame skeleton snippet of 38 upper-body joints (114 dimensions per frame). The input is then split into its body stream consisting of 8 joints (neck, head, shoulders, elbows, wrists) and its hand stream (3 joints per finger \& 30 in total). Each of these streams is then independently encoded by a 1D convolutional encoder consisting of Conv1D layers, and combined with Group Normalization and LeakyReLU activations. Each Conv1D layer uses kernel size 4, stride 2, and padding 1. This yields a downsampling of 8 frames into 1 latent vector of dimension 128 per stream. Each latent sequence is independently quantised by a Residual Vector Quantiser. Our final configuration uses three residual stages with codebook sizes (128, 128, 128) for the body stream and three with sizes (128, 64, 32) for the hand stream. All codebooks are first initialized via k-means clustering on encoder outputs from one training batch and updated during training using EMA updates with decay $\mu=0.99$ to prevent codebook collapse. Codebook entries whose EMA count falls below the reset threshold of 1.0 are replaced with randomly sampled encoder outputs from the current batch, preventing inactive codes from remaining unused. The quantized body and hand latents are concatenated along the feature dimension and passed to a shared transposed convolutional decoder, which mirrors the encoder. It consists of two ConvTranspose1D layers using kernel size 4, stride 2, and padding 1, with Group Normalisation and Leaky ReLU, followed by a final ConvTranspose1D layer. The decoder upsamples the concatenated latents by a factor of 8 to reconstruct the full 38-joint skeleton sequence across the 8 frames. 

The model is trained on 8-frame chunks using a stream-decoupled reconstruction objective. It is trained with the Adam optimizer using a learning rate of $10^{-3}$. Early stopping is applied using validation loss with a patience of 5 epochs. We use dataloaders with a clip-level batch size of 8. Two separate decoder forward passes are performed. In the first, the hand latent is detached via a stop-gradient operator so that body reconstruction error flows gradients only through the body encoder and body quantizer. In the second, the body latent is detached so that hand reconstruction error flows only through the hand stream. This decoupling prevents the numerically dominant hand joints (30 of 38) from overwhelming the body stream's gradient signal. The quantization loss is a commitment loss only, as codebook updates are handled implicitly by EMA. The total loss objective is:

\begin{equation}
    \mathcal{L} = \mathcal{L}_\text{body} + \mathcal{L}_\text{hand} + \mathcal{L}_\text{vq}.
\end{equation}

We report the MPJPE values in normalized coordinate space in Table \ref{tab:rvqvae_ablation} for different codebook configurations. In addition to codebook configurations, we also report MPJPE values with different temporal downsampling factors. Smaller factors preserve more temporal detail but produce longer token sequences, while larger factors yield more compact tokenizations at the cost of reconstruction quality. Table~\ref{tab:downsampling_ablation} shows that reconstruction error increases as more frames are compressed into a single token. We use a downsampling factor of 8 in the main experiments as a trade-off between reconstruction fidelity, compact sequence length and verbalization reliability: longer primitives are more difficult to summarize with a concise textual description.

\begin{table}[h]
\centering
\resizebox{\linewidth}{!}{%
\begin{tabular}{ccccccc}
\hline
\textbf{Body CBs} & \textbf{Hand CBs} & \textbf{Dim} &
\textbf{MPJPE (Body)} & \textbf{MPJPE (Hand)} & \textbf{MPJPE (All)} & \textbf{Jitter} \\
\hline
$(128)$             & $(64)$           & 64  & 0.0403 & 0.0878 & 0.0777 & 0.0065 \\
$(128)$             & $(128)$           & 128  & 0.0386 & 0.0851 & 0.0762 & 0.0064 \\
$(128)$             & $(512)$           & 128  & 0.0357 & 0.0723 & 0.0646 & 0.0064 \\
$(128, 128)$        & $(64, 4, 2)$      & 128  & 0.0314 & 0.0658 & 0.0585 & 0.0065 \\
$(128, 128)$        & $(128, 64)$        & 128  & 0.0300 & 0.0602 & 0.0538 & 0.0063 \\
$(128, 128, 128)$   & $(128, 64, 32)$   & 128 & \textbf{0.0253} & \textbf{0.0493} & \textbf{0.0442} & \textbf{0.0060} \\
\hline
\end{tabular}%
}
\caption[RVQ-VAE architectural ablation experiments]{Architectural ablation over selected codebook configurations and latent dimensions. Body and hand codebook entries denote the size of each residual quantisation stage. The latent downsampling factor was fixed at 8. MPJPE is reported separately for outer body joints, hand joints, and all joints combined. The final row corresponds to the selected model configuration. All metrics are reported on the combined test dataset.}
\label{tab:rvqvae_ablation}
\end{table}

\begin{table}[h]
\centering
\begin{tabular}{cc}
\toprule
\textbf{Downsampling Factor} & \textbf{Test MPJPE (All)} \\
\hline
4 frames  & 0.0397 \\
8 frames  & 0.0442 \\
16 frames & 0.0529 \\
\bottomrule
\end{tabular}
\caption{MPJPE for three temporal downsampling factors, keeping all other hyperparameters fixed. Larger factors compress more frames into a single codebook assignment, resulting in higher reconstruction error. All metrics are reported on the combined test dataset.}
\label{tab:downsampling_ablation}
\end{table}


\subsection{Rule-based Motion Primitive Verbalization}
\label{appx:verbalisation}

Each motion token is defined by the residual codebook indices of the body and hand streams. To verbalize each of these tokens, we first reconstruct the corresponding 8-frame skeleton sequence from its body and hand codebook indices, and then apply deterministic geometric rules to extract interpretable physical attributes. Doing this across all body and hand tokens yields a lookup dictionary for each combination of body and hand token indices. 
We construct two variants of this lookup dictionary. The first enumerates all possible joint body-hand token combinations, i.e., all combinations across the body codebooks \((128 \times 128 \times 128)\) and hand codebooks \((128 \times 64 \times 32)\). This produces descriptions for every complete body-hand primitive, but results in a very large dictionary. The second variant constructs separate stream-wise dictionaries: for the body dictionary, all body token combinations are decoded while the hand latent is held fixed at its mean value; for the hand dictionary, all hand token combinations are decoded while the body latent is held fixed at its mean value. We use this stream-wise lookup dictionary in our experiments because it preserves the same body and hand verbalization procedure while requiring only \((128^3) + (128 \times 64 \times 32)\) entries instead of \((128^3) \times (128 \times 64 \times 32)\) entries.

\subsubsection{Body-stream Attributes}
\label{body_stream_attributes}

All spatial attributes are computed in a body-normalized coordinate frame. For each frame, we compute the shoulder width $w_\text{shoulder}$ as the Euclidean distance between the left and right shoulder joints,
and use the head-to-neck distance \(d_{\text{head}}\) as a vertical scale reference. The shoulder, chest, torso, and waist levels are estimated as
\begin{equation}
y_{\text{shoulder}} =
\frac{y_{\text{left shoulder}} + y_{\text{right shoulder}}}{2},
\qquad
y_{\text{chest}} =
y_{\text{shoulder}} - 0.5 d_{\text{head}},
\end{equation}
\begin{equation}
y_{\text{waist}} =
y_{\text{shoulder}} - 1.5 d_{\text{head}},
\qquad
y_{\text{torso}} =
\frac{y_{\text{chest}} + y_{\text{waist}}}{2}.
\end{equation}
These landmarks are used to classify wrist height of the skeleton. For each wrist, we extract vertical level, horizontal placement, depth, elbow bend, reach, and motion direction.

\paragraph{Wrist vertical level:}
Wrist height is assigned to one of six classes:
\begin{table}[h]
\centering
\begin{tabular}{ll}
\hline
\textbf{Label} & \textbf{Condition} \\
\hline
above-head &
\(y_{\text{wrist}} > y_{\text{head}}\) \\
shoulder-level &
\(y_{\text{wrist}} > y_{\text{shoulder}}\) \\
chest-level &
\(y_{\text{wrist}} > y_{\text{chest}}\) \\
torso-level &
\(y_{\text{wrist}} > y_{\text{torso}}\) \\
waist-level &
\(y_{\text{wrist}} > y_{\text{waist}}\) \\
below-waist &
otherwise \\
\hline
\end{tabular}
\caption{Wrist vertical level classification thresholds used in rule-based primitive verbalization.}
\label{tab:wrist-level-thresholds}
\end{table}

\paragraph{Wrist horizontal position:}
Horizontal placement is computed relative to the neck-centered body midline and the ipsilateral shoulder. The outward displacement is
\begin{equation}
\delta_{\text{outward}} =
\begin{cases}
x_{\text{shoulder}} - x_{\text{wrist}}, & \text{left arm}, \\
x_{\text{wrist}} - x_{\text{shoulder}}, & \text{right arm}.
\end{cases}
\end{equation}
A wrist is labeled \textit{crossed-inward} if it crosses the body center by more than \(0.05\) units, \textit{extended-outward} if
\(\delta_{\text{outward}} > 0.4 w_{\text{shoulder}}\), \textit{torso-side} if it is at or beyond the shoulder without passing the extended threshold, and \textit{body-centre} otherwise.

\paragraph{Wrist depth:}
Depth is classified from the wrist \(z\)-coordinate as \textit{in-front-of-torso} if
\[
z_{\text{wrist}} < -0.15,
\]
\textit{at-torso} if
\[
z_{\text{wrist}} < 0.05,
\]
and \textit{behind-torso} otherwise.

\paragraph{Elbow bend:}
The elbow angle is computed from the upper-arm and forearm vectors:
\begin{equation}
\theta_{\text{elbow}} =
\arccos\left(
\operatorname{clip}\left(
\frac{-(p_{\text{elbow}}-p_{\text{shoulder}})}
{\|p_{\text{elbow}}-p_{\text{shoulder}}\|}
\cdot
\frac{p_{\text{wrist}}-p_{\text{elbow}}}
{\|p_{\text{wrist}}-p_{\text{elbow}}\|},
-1,1
\right)
\right).
\end{equation}
It is categorized as \textit{sharply-bent} \((<45^\circ)\), \textit{bent} \((<90^\circ)\), \textit{slightly-bent} \((<135^\circ)\), or \textit{straight} \((\geq 135^\circ)\).

\paragraph{Arm reach:}
Reach is the ratio between wrist-to-shoulder distance and total arm length:
\begin{equation}
\rho =
\frac{
\|p_{\text{wrist}} - p_{\text{shoulder}}\|
}{
\|p_{\text{elbow}} - p_{\text{shoulder}}\|
+
\|p_{\text{wrist}} - p_{\text{elbow}}\|
+
\epsilon
}.
\end{equation}
It is labeled \textit{near-body} if \(\rho < 0.4\), \textit{mid-reach} if \(0.4 \leq \rho \leq 0.7\), and \textit{extended} if \(\rho > 0.7\).

\paragraph{Arm motion direction:}
Wrist displacement across the 8-frame window is used to detect motion. If both total displacement and maximum single-frame displacement are below \(0.03\), the arm is labeled \textit{held}. Otherwise, motion is decomposed into vertical, horizontal, and depth components: \textit{rising} or \textit{lowering} from \(\Delta y\), \textit{moving-inwards} or \textit{moving-outwards} from \(\Delta x\) relative to body side, and \textit{moving-forward} or \textit{moving-backward} from \(\Delta z\). When motion occurs on multiple axes, the direction labels are concatenated.

\subsubsection{Hand-stream Attributes}

For each hand, we extract palm orientation and hand shape.

\paragraph{Palm orientation:}
The palm normal is estimated using the cross product of two vectors on the palm plane: the vector from pinky base to index base, and the vector from wrist to palm center, which is the mean
position of the four non-thumb finger base joints:
\begin{equation}
n_{\text{palm}} =
\frac{
(p_{\text{index base}} - p_{\text{pinky base}})
\times
(p_{\text{palm centre}} - p_{\text{wrist}})
}{
\left\|
(p_{\text{index base}} - p_{\text{pinky base}})
\times
(p_{\text{palm centre}} - p_{\text{wrist}})
\right\|
}.
\end{equation}
For the right hand, the normal is negated to account for handedness. The dominant axis of the normal determines the label: \(x\)-dominant gives \textit{facing-left} or \textit{facing-right}, \(y\)-dominant gives \textit{facing-up} or \textit{facing-down}, and \(z\)-dominant gives \textit{facing-speaker} or \textit{facing-away}. Left/right labels are remapped into body-relative \textit{facing-inward} or \textit{facing-outward} labels depending on the hand side.

\paragraph{Finger curl and hand shape:}
For each finger, curl is measured as the angle between the base-to-mid and mid-to-tip segments:
\begin{equation}
\theta^{(f)}_{\text{curl}} =
\arccos\left(
\operatorname{clip}\left(
\hat{v}^{(f)}_{\text{base-mid}}
\cdot
\hat{v}^{(f)}_{\text{mid-tip}},
-1,1
\right)
\right).
\end{equation}
The mean curl of the four non-thumb fingers is used for classification. The pipeline first checks for \textit{index-pointing}, defined as index curl below \(25^\circ\) and mean curl of the other three fingers above \(40^\circ\). Remaining cases are classified as \textit{open-flat} \((\bar{\theta}<20^\circ)\), \textit{open-relaxed} \((\bar{\theta}<35^\circ)\), \textit{curled} \((\bar{\theta}<55^\circ)\), or \textit{fist} \((\bar{\theta}\geq55^\circ)\).

All of the above extracted attributes are mapped to text using deterministic templates. Each primitive produces a body description or a hand description, depending on the stream. Example descriptions of a body and hand primitive are shown below:
\begin{quote}
\small
"body": "Left wrist held at shoulder level (vertical), in front of
torso (depth) and torso side (horizontal); elbow bent, reach mid-reach; Right wrist held waist level
(vertical), in front of torso (depth) and body center (horizontal); elbow bent, reach mid-reach."

"hands": "Left palm facing outward, hand shape changing from curled to open relaxed; Right
palm facing inward, hand shape curled, held."
\end{quote}
During inference for every gesture sample, the primitive-level descriptions are concatenated chronologically, so a 24-frame gesture yields three 8-frame verbalizations each for the body and hand streams.

\subsubsection{Temporal aggregation of attributes}

All discrete body and hand attributes are first classified independently for each frame in the reconstructed 8-frame primitive. These include wrist level, depth, horizontal placement, elbow bend, reach, palm orientation, and hand shape. To obtain a sequence-level description, we use the middle-frame label as the representative state when the attribute remains unchanged. To detect changes in attributes across the sequence, we compare the labels assigned to the first and last frames. If they differ, our template reports an explicit transition, e.g., "hand shape changing from curled to open relaxed" or "palm rotating from facing inward to facing outward"; otherwise, the attribute is described using its middle-frame label. This aggregation is applied separately to the left and right hands. If both hands share the same orientation, hand shape, and transition pattern, they are collapsed into a single bimanual description to avoid redundancy. Otherwise, the hands are verbalized separately.

\subsubsection{Attribute Extraction Validation}

The generated physical descriptions receive human Pose Scores of 3.75/5 on TED and 3.77/5 on BEAT2, with strong LLM–human agreement ($\rho = 0.887$ and $0.942$; Sec.~\ref{sec:quality_evaluation}). To examine possible confusions, we further audited 100 BEAT2 clips against free-text annotator descriptions. Most disagreements involved palm orientations near adjacent geometric thresholds or symbolic hand configurations (e.g., an OK sign) represented by broader hand-shape categories, reflecting the deliberately coarse granularity of the attribute vocabulary rather than systematic confusion between substantially different gestures. A raw-input comparison further shows that intent scores are unchanged whether attributes are extracted from raw joint positions or RVQ-VAE-reconstructed positions; tokenized inputs yield slightly higher LLM-judge Pose Scores (TED: 3.4 vs. 3.2 on 231 samples; BEAT2: 3.2 vs. 3.1 on 100 samples), so tokenization does not hurt intent quality and slightly improves physical-form descriptions.


\subsection{LLM prompts for description generation and evaluation}

\subsubsection{Structured Reasoning-Based Prompt for Description Generation}
\label{prompt:generation}

\begin{tcolorbox}[
  breakable,
  enhanced,
  fonttitle=\bfseries,
  coltitle=black,
  colbacktitle=white,
  fontupper=\footnotesize,
  colback=white,
  colframe=gray!50,
  left=6pt, right=6pt, top=6pt, bottom=6pt,
  boxsep=0pt,
  before skip=8pt,
  after skip=8pt
]

\setlength{\parskip}{1pt}
\setlength{\parindent}{0pt}

You are an expert in gesture analysis and nonverbal communication. Your task is to generate a semantic description of a gesture performed by a speaker during a talk. This description should capture both the physical motion and the communicative intent of the gesture.

\smallskip
\textbf{Input Data}

You will receive:
\begin{enumerate}
  \setlength{\itemsep}{0pt}
  \setlength{\topsep}{1pt}
  \item \textbf{gesture\_text} --- the phrase being spoken during the gesture
  \item \textbf{text\_context} --- additional surrounding context of the utterance
  \item \textbf{verbalised\_motion\_sequence} --- a sequence of body and hand descriptions across time. Each entry represents 8 frames of motion and includes:
  \begin{itemize}
    \setlength{\itemsep}{0pt}
    \setlength{\topsep}{1pt}
    \item \texttt{body}: wrist positions (vertical level, depth, horizontal placement), elbow bend, and reach
    \item \texttt{hands}: hand shape, palm orientation, and relative hand movements
  \end{itemize}
\end{enumerate}

\smallskip
\textbf{Important Notes About the Data}

\begin{itemize}
  \setlength{\itemsep}{0pt}
  \setlength{\topsep}{1pt}
  \item The motion capture data is imperfect and lacks fine-grained detail, especially for fingers and hand shapes.
  \item The \texttt{hands} verbalisations are coarse approximations---do \textbf{not} over-anchor on specific finger configurations.
  \item Do \textbf{not} hallucinate fine finger articulation unless the speech context strongly implies a symbolic gesture, e.g., pointing or counting.
  \item Focus on the overall motion trajectory, spatial relationships, and timing patterns, which are more reliably captured.
  \item \textbf{Critical:} The verbalisations describe \textbf{both} hands even when only \textbf{one} hand performs the meaningful gesture. You must actively determine handedness---do \textbf{not} default to "both hands".
\end{itemize}

\smallskip
\textbf{Your Task}

Generate a single, concise semantic description, 1--2 sentences, that describes:
\begin{enumerate}
  \setlength{\itemsep}{0pt}
  \setlength{\topsep}{1pt}
  \item \textbf{The physical motion} --- what the hands/arms do in space
  \item \textbf{The communicative intent} --- what meaning or emphasis the gesture conveys in context
\end{enumerate}

\smallskip
\textbf{Stage 0: Determine Handedness (critical)}

The verbalisations always describe both hands, but often only \textbf{one} hand performs the meaningful gesture.

Privately consider, in order:

\textbf{A) Speech-based triggers for one-hand gestures}
\begin{itemize}
  \setlength{\itemsep}{0pt}
  \setlength{\topsep}{1pt}
  \item Numbers or quantity words, e.g., "one", "two", "first", "second", "single", "only" $\rightarrow$ likely one-hand pointing/counting
  \item Deixis, e.g., "this", "that", "here", "there", "it" $\rightarrow$ likely one-hand pointing
  \item Contrast or alternatives, e.g., "but", "however", "or", "otherwise", "instead" $\rightarrow$ likely one-hand presenting an alternative
  \item Very short, emphatic phrase, 1--3 words $\rightarrow$ likely one-hand beat/emphasis
  \item Refers to a single entity being singled out $\rightarrow$ likely one-hand
\end{itemize}

\textbf{B) Speech-based triggers for both-hands gestures}
\begin{itemize}
  \setlength{\itemsep}{0pt}
  \setlength{\topsep}{1pt}
  \item Size, extent, or magnitude, e.g., "big", "huge", "large", "small", "wide", "all", "everything", "whole" $\rightarrow$ likely both hands depicting size
  \item Relationship, comparison, or two-sided concept, e.g., "between", "from X to Y", "versus", "and", "together" $\rightarrow$ likely both hands
  \item Containment or framing, e.g., "within", "inside", "around", "the whole" $\rightarrow$ likely both hands framing
  \item Inclusive self/group reference, e.g., "we", "us", "our", "ourselves" $\rightarrow$ could be both hands inward motion
  \item Listing multiple items $\rightarrow$ could be both hands alternating
\end{itemize}

\textbf{C) Motion sequence-based signals}
\begin{itemize}
  \setlength{\itemsep}{0pt}
  \setlength{\topsep}{1pt}
  \item One hand static while the other shows motion verbs $\rightarrow$ the moving hand is active, one-hand gesture
  \item Symmetric, coordinated motion $\rightarrow$ both hands
  \item One hand near-body/rest-like while the other is extended $\rightarrow$ one-hand gesture, using the extended hand
  \item Opposing directions or clear bimanual coordination, e.g., apart, together, framing $\rightarrow$ both hands
\end{itemize}

\textbf{D) Make a firm commitment}
\begin{itemize}
  \setlength{\itemsep}{0pt}
  \setlength{\topsep}{1pt}
  \item If speech suggests one hand and verbalisations show asymmetry $\rightarrow$ definitely one-hand
  \item If speech suggests one hand but verbalisations look symmetric $\rightarrow$ still likely one-hand; trust speech over verbalisations
  \item If speech suggests both hands and verbalisations show coordination $\rightarrow$ definitely both hands
  \item If unclear, prefer one-hand if the gesture could reasonably be performed with one hand
\end{itemize}

Privately commit to: \textbf{Handedness = one hand / both hands}. If one-hand, decide which hand is more active, but prefer "one hand" without specifying left/right unless clearly evident.

\smallskip
\textbf{Stage 1: Infer Communicative Intent (from text)}

Privately decide:
\begin{itemize}
  \setlength{\itemsep}{0pt}
  \setlength{\topsep}{1pt}
  \item What the speaker is communicating and what they might want to emphasize or illustrate
  \item The best intent category, choosing one: emphasis, listing, enumeration, contrast, uncertainty, self-reference, other references, discourse, temporal progression/reference, relativity, emotion, negation, quantification or symbolic depiction
  \item The referent: self / other / concrete object / abstract idea / imagined space / audience / number or quantity
\end{itemize}

\smallskip
\textbf{Stage 2: Map Motion from Verbalisations}

Privately decide:
\begin{itemize}
  \setlength{\itemsep}{0pt}
  \setlength{\topsep}{1pt}
  \item Level: waist / chest / shoulder / head; use "chest level" by default unless clearly waist/shoulder
  \item Primary motion path: up / down / inward / outward / forward / backward / lateral sweep / arc / circular / held/static / pointing
  \item If both hands: relation, e.g., together / apart / framing / mirroring / alternating
  \item Palm orientation and hand shape only if clearly supported
\end{itemize}

For symbolic gestures, such as pointing, counting, or indicating:
\begin{itemize}
  \setlength{\itemsep}{0pt}
  \setlength{\topsep}{1pt}
  \item If language implies singularity, e.g., "one", infer index-finger pointing
  \item If language includes numbers, infer finger counting configuration
  \item If language includes "this", "that", "here", or "there", infer pointing
\end{itemize}

\smallskip
\textbf{Stage 3: Draft and Verify}

Privately:
\begin{enumerate}
  \setlength{\itemsep}{0pt}
  \setlength{\topsep}{1pt}
  \item Draft a 1--2 sentence description combining handedness, motion, and intent
  \item Verify correct handedness, level, primary motion, and intent
  \item If any check fails, especially handedness, revise
  \item Output only the final verified description
\end{enumerate}

\smallskip
\textbf{Style Requirements}

\begin{itemize}
  \setlength{\itemsep}{0pt}
  \setlength{\topsep}{1pt}
  \item Begin with: "One hand\ldots", "Both hands\ldots", or rarely "The right/left hand\ldots"
  \item For pointing gestures: "One hand points forward with index finger extended\ldots" or "One hand points using the forefinger\ldots"
  \item Use concrete motion verbs: points, moves, sweeps, pushes, pulls, rises, extends, rotates, spreads
  \item Use "chest level" as the default level, not "torso level", unless clearly waist or shoulder
  \item Include palm/hand orientation only if supported, e.g., "with palm facing up"
  \item End with communicative function: "to emphasize\ldots", "to indicate\ldots", "to depict\ldots", "to convey\ldots", "to refer to\ldots"
  \item Keep it concise: 15--35 words, 1--2 sentences
  \item Use present tense
  \item Do \textbf{not} mention frames, chunks, VQVAE, codes, verbalisations, or any meta-commentary
\end{itemize}

\smallskip
\textbf{Output Format}

Output only the final semantic description. No reasoning, no tags, no explanations, no preamble.

\smallskip
Now privately reason through all stages, starting with handedness, verify your draft, and output only the final semantic description for this input.

\end{tcolorbox}

\subsubsection{LLM-as-a-Judge evaluation prompt for descriptions}
\label{prompt:evaluation}

\begin{tcolorbox}[
  breakable,
  enhanced,
  fonttitle=\bfseries,
  coltitle=black,
  colbacktitle=white,
  fontupper=\footnotesize,
  colback=white,
  colframe=gray!50,
  left=6pt, right=6pt, top=6pt, bottom=6pt,
  boxsep=0pt,
  before skip=8pt,
  after skip=8pt
]

\setlength{\parskip}{1pt}
\setlength{\parindent}{0pt}

You evaluate gesture descriptions. Compare the generated description to the gold reference. You will receive two short texts:
\begin{itemize}
    \setlength{\itemsep}{0pt}
    \setlength{\topsep}{1pt}
    \item \textbf{REFERENCE (Gold)} --- the ground-truth semantic description of a gesture.
    \item \textbf{CANDIDATE (Generated)} --- a model-generated semantic description of the same gesture.
\end{itemize}

\smallskip
\textbf{Core Principle:} Focus on whether the descriptions capture the same gesture type, not exact wording.

\smallskip
\textbf{PoseScore (1--5): Physical Gesture Similarity}
\begin{itemize}
    \setlength{\itemsep}{0pt}
    \setlength{\topsep}{1pt}
    \item \textbf{5 = Same gesture:} same body parts, same upper-body region, and same motion category.
    \item \textbf{4 = Very similar:} same body parts and region, with only minor wording differences, e.g., "spreads apart" vs. "moves outward".
    \item \textbf{3 = Partially similar:} same body parts, but different motion or level while still plausible.
    \item \textbf{2 = Related but different:} different number of hands or noticeably different motion type.
    \item \textbf{1 = Completely different:} opposite configuration or unrelated gesture.
\end{itemize}

\smallskip
\textbf{IntentScore (1--5): Communicative Function}
\begin{itemize}
    \setlength{\itemsep}{0pt}
    \setlength{\topsep}{1pt}
    \item \textbf{5 = Same intent:} emphasis, pointing, size depiction, self-reference, etc.
    \item \textbf{4 = Very similar intent}
    \item \textbf{3 = Plausible alternative interpretation}
    \item \textbf{2 = Loosely related}
    \item \textbf{1 = Contradictory}
\end{itemize}

\smallskip
\textbf{Equivalences:} Treat the following as equivalent:
\begin{itemize}
    \setlength{\itemsep}{0pt}
    \setlength{\topsep}{1pt}
    \item "chest level" = "torso level" = "in front of body"
    \item "moves apart" = "spreads" = "separates" = "outward"
    \item "pulls inward" = "moves toward body" = "brings together"
    \item "emphasize" = "mark" = "stress" = "underscore"
\end{itemize}

\smallskip
\textbf{Important:} Be lenient on motion direction differences if gesture type matches. "One hand" vs. "Both hands" is a significant difference, with a score of 2--3 maximum. Focus on semantic similarity, not lexical matching.

\smallskip
\textbf{Output:} Output exactly three lines:
\begin{quote}
\ttfamily\footnotesize
PoseScore: [1--5]\\
IntentScore: [1--5]\\
AverageScore: [1--5]
\end{quote}

No explanation.

\end{tcolorbox}

\subsection{Ablations: Semantic motion anchor Generation} 
\label{appx:semantic-anchor-generation}
To assess prompt sensitivity, we compare four ways of converting the same RVQ-VAE motion narrative and transcript into a semantic motion anchor: a naive zero-shot prompt, an in-context prompt with gesture examples, a chain-of-thought prompt that separates handedness, motion, and intent, and our structured reasoning prompt, which further enforces handedness decisions, motion-intent consistency, and constraints on spatial and hand-shape evidence.

We evaluate all variants using the LLM-as-judge setup from Section~\ref{sec:quality_evaluation} on 231 TED and 100 BEAT semantic gesture samples.

\begin{table}[h]
\centering
\small
\begin{tabular}{lcccccc}
\toprule
& \multicolumn{2}{c}{TED ($N=231$)}
& \multicolumn{2}{c}{BEAT2 ($N=100$)}
& \multicolumn{2}{c}{Weighted Avg.} \\
\cmidrule(lr){2-3}
\cmidrule(lr){4-5}
\cmidrule(lr){6-7}
Prompt & Pose & Intent & Pose & Intent & Pose & Intent \\
\hline
Naive prompt & 3.0 & 4.1 & 3.3 & \textbf{4.4} & 3.1 & 4.2 \\
In-context learning prompt & 3.1 & 4.1 & 3.3 & 4.4 & 3.2 & 4.1 \\
Chain-of-thought prompt & 3.0 & \textbf{4.2} & \textbf{3.4} & 4.4 & 3.1 & 4.2 \\
Structured reasoning prompt & \textbf{3.4} & 4.1 & 3.2 & 4.3 & \textbf{3.3} & \textbf{4.2} \\
\bottomrule
\end{tabular}
\caption{Prompt sensitivity analysis for semantic motion anchor generation using token-based motion narratives. Scores are LLM-as-a-judge ratings on a 1--5 scale.}
\label{tab:prompt-sensitivity}
\end{table}

The structured reasoning prompt gives the best overall pose score, while Intent Score remains high across all prompt variants. This suggests that intent is often recoverable from transcript context, whereas accurate physical-form description depends more strongly on how the prompt guides the model to use the motion narrative. We therefore use the structured reasoning prompt for all downstream experiments.

\subsection{Cross-modal training and inference: Implementation Details}
\label{appx:cross-modal-implementation-details}
The motion encoder $f_\mathrm{mot}$ is a 2-layer, 4-head Transformer with hidden dimension 256 and maximum sequence length 1024. All projection MLPs map into a shared 512-dimensional retrieval space via LayerNorm $\to$ Linear $\to$ GELU $\to$ Dropout(0.1) $\to$ Linear, with L2-normalized outputs. The contrastive temperature is initialized at $\tau{=}0.07$ and learned during training. We train with AdamW ($\text{lr}{=}5{\times}10^{-5}$, weight decay $10^{-4}$, gradient clipping 1.0, constant schedule), batch size 512, for up to 40 epochs with early stopping (patience 10). The model is first warmed up on the transcript--motion objective $\mathcal{L}_\mathrm{TM}$ alone. For the full objective (Eq.~\ref{eq:main}), we set $\lambda_s{=}0.15$, $\lambda_p{=}0.03$, and $\lambda_b{=}0.02$. All models are trained on a single H100 GPU.

\paragraph{Comparison Approaches.}
All baseline retrieval models are evaluated using identical data splits, motion encoders, projection heads, batch sizes, and evaluation protocols unless otherwise noted. To ensure fair comparison, our plain text-motion baseline utilizes the exact same retrieval backbone but is trained exclusively with the standard transcript-motion contrastive objective $\mathcal{L}_{tm}$, omitting the semantic motion anchor framework entirely. GestureDiffuCLIP replaces our Qwen3-Embedding-8B text encoder with a frozen CLIP ViT-B/32 (512-dim, no instruction prompts) and trains with a plain symmetric InfoNCE loss between transcripts and motion, with no negative-handling or anchor supervision. TMR uses the same Qwen3-Embedding-8B encoder as ours but trains with transcript-motion InfoNCE augmented by false-negative filtering, masking within-batch pairs whose transcript cosine similarity exceeds 0.9 following \citep{petrovich2023tmr}, JEGAL (text-only) also uses Qwen and replaces hard negatives with soft positive targets: pairs with transcript cosine similarity above 0.85 are assigned a partial positive weight of 0.5, adapting JEGAL's global phrase contrastive objective to our setting. The only varying factor across methods is the training objective, making the comparison a direct test of loss design.

\subsection{Marginal Sensitivity of Auxiliary Loss Weights}
\label{app:marginal_sensitivity}

\begin{figure}[h]
    \centering
    \includegraphics[width=\linewidth]{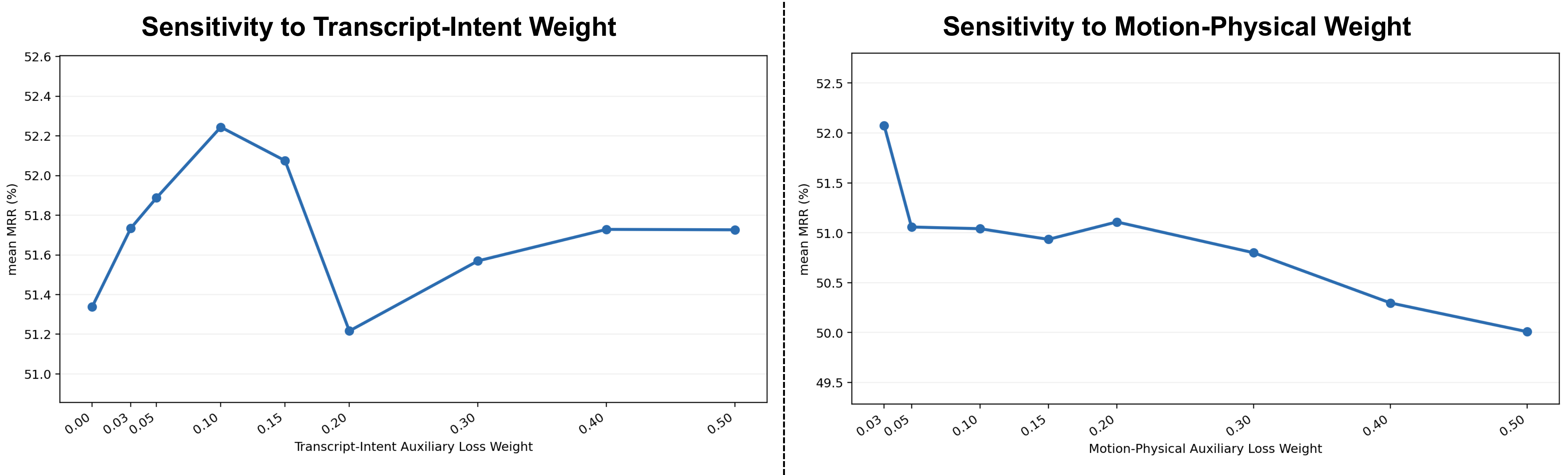}
    \caption{Marginal sensitivity of auxiliary loss weights on mean MRR (\%) (text -> Motion and Motion -> Text).}
    \label{fig:marginal_sensitivity}
\end{figure}

Figure~\ref{fig:marginal_sensitivity} complements the joint sensitivity heatmap in Figure~\ref{fig:weight_sensitivity} 
by showing how mean MRR varies when each auxiliary weight is swept independently while the other is held fixed. Specifically, $\lambda_p$ is fixed at $0.03$ for the 
transcript-intent sweep, and $\lambda_s$ is fixed at $0.15$ for the motion-physical sweep, 
both corresponding to the near-optimal region identified in the joint analysis.

The two sweeps together reinforce the asymmetric behaviour observed in the joint 
heatmap. The transcript-intent branch tolerates a moderate range of $\lambda_s$ values 
without strong degradation, whereas even small increases in $\lambda_p$ beyond the 
near-zero optimum consistently hurt retrieval performance. This further supports the 
recommendation to keep $\lambda_p$ small while allowing $\lambda_s$ to have more flexibility.

\begin{figure}[h]
    \centering
    \includegraphics[width=0.55\linewidth]{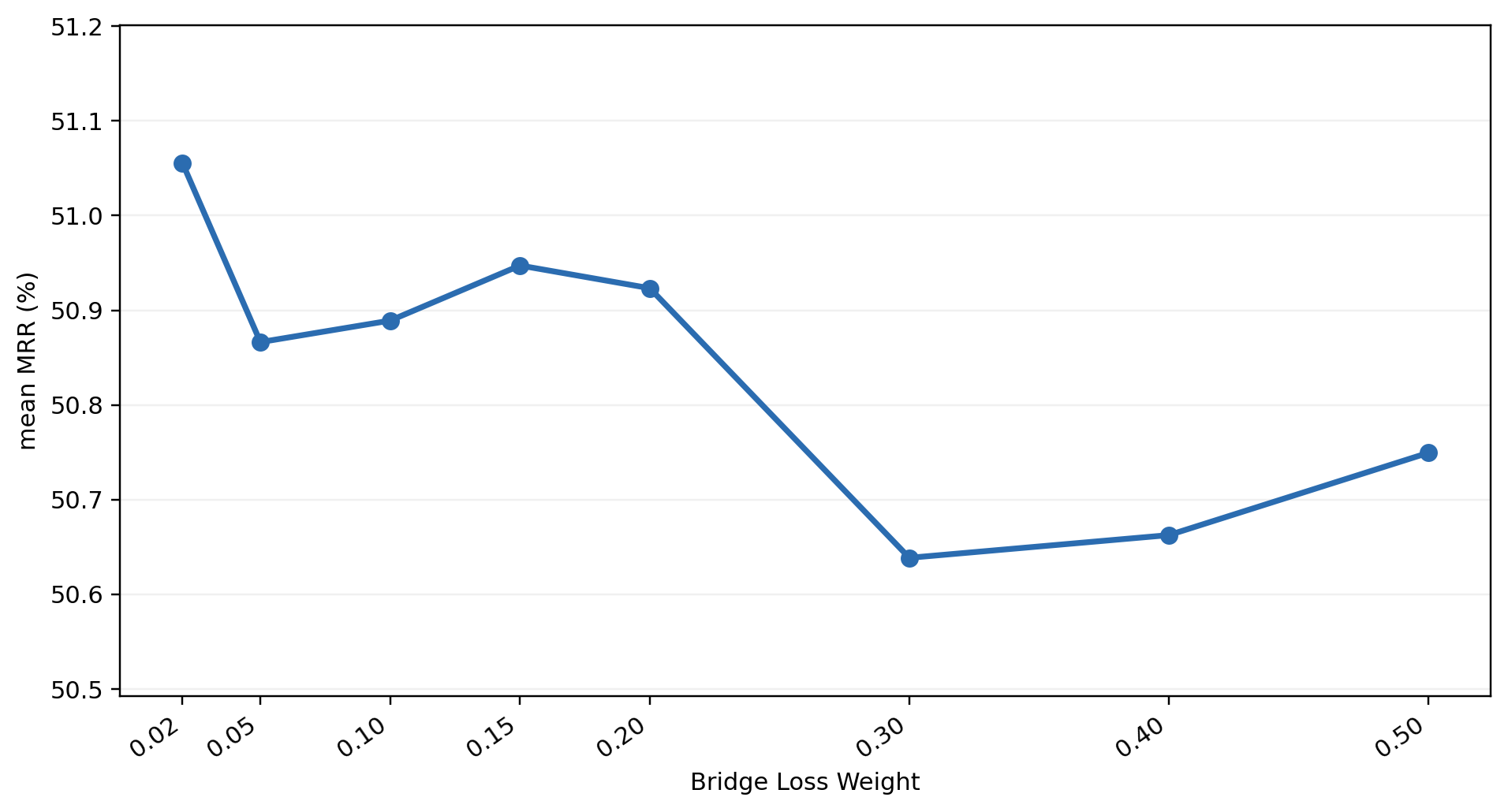} 
    \caption{Sensitivity of mean MRR (\%) to the bridge loss weight ($\lambda_b$), with $\lambda_p$ and $\lambda_s$ fixed.}
    \label{fig:bridge_loss_sensitivity}
\end{figure}

In addition, Figure~\ref{fig:bridge_loss_sensitivity} extends this analysis by sweeping the bridge loss weight, $\lambda_b$, while holding the other auxiliary weights fixed at their near-optimal values ($\lambda_p = 0.03$ and $\lambda_s = 0.15$). The results demonstrate that the bridge loss weight should be kept very small, with performance peaking at $\lambda_b = 0.02$. The primary role of $\mathcal{L}_{br}$ is to regularize the shared anchor space at a low weight, preventing the physical-form and communicative-intent representations from drifting apart. Increasing $\lambda_b$ beyond this point forces these representations to become overly constrained, which actively degrades overall retrieval performance. 

\subsection{Decomposing the Semantic Motion Anchor into Physical-Form and Intent Components}
\label{app:anchor_split}

The unified semantic motion anchor $a$ produced by the structured reasoning prompt encodes both gesture form and communicative intent in a single 1--2 sentence description. To enable modality-matched contrastive supervision, we decompose each anchor into its physical-form component $a^{phys}$ and communicative-intent component $a^{int}$ via a separate zero-shot prompt to Qwen3-8B that emits a JSON object with two fields. The decomposition is run offline once over the full training set of the BEAT2 dataset. Decoding is greedy with a fixed token budget of $160$; the system and user prompts are reproduced verbatim below.

\begin{tcolorbox}[
  breakable,
  enhanced,
  fonttitle=\bfseries,
  coltitle=black,
  colbacktitle=white,
  fontupper=\footnotesize,
  colback=white,
  colframe=gray!50,
  left=6pt, right=6pt, top=6pt, bottom=6pt,
  boxsep=0pt,
  before skip=8pt,
  after skip=8pt
]

\setlength{\parskip}{1pt}
\setlength{\parindent}{0pt}

\textbf{System prompt}

\smallskip
You split gesture descriptions into exactly two aspects. Return JSON with exactly these keys:
\begin{itemize}
  \setlength{\itemsep}{0pt}
  \setlength{\topsep}{1pt}
  \item \texttt{physical\_motion}
  \item \texttt{semantic\_intent}
\end{itemize}

\smallskip
\textbf{Rules}
\begin{itemize}
  \setlength{\itemsep}{0pt}
  \setlength{\topsep}{1pt}
  \item \texttt{physical\_motion}: only the visible body movement, hand shape, position, direction, timing, posture, and spatial form.
  \item \texttt{semantic\_intent}: only the communicative meaning, emphasis, discourse function, emotion, or concept conveyed by the gesture.
  \item Keep both fields concise but faithful.
  \item Do not repeat the same content in both fields.
  \item If semantic meaning is unclear, set \texttt{semantic\_intent} to an empty string.
  \item Output valid JSON only. No markdown.
\end{itemize}

\smallskip
\textbf{User prompt template}

\smallskip
Split this gesture description into physical motion and semantic intent. Return only a JSON object like \texttt{\{"physical\_motion":"...","semantic\_intent":"..."\}}.

\smallskip
Description: \texttt{<unified anchor>}

\end{tcolorbox}

The two fields are subsequently embedded independently by the frozen Qwen3-Embedding-8B text encoder $g_{\text{text}}$ and projected through the shared anchor projector $\pi_{an}$ to obtain $z_p$ and $z_s$.

\subsection{Random Anchor Baseline: Implementation Details}
\label{app:random_anchor}

In the Random Anchor baseline, the text encoder outputs $g_{\text{text}}(a^{phys}_i)$ and $g_{\text{text}}(a^{int}_i)$ are each replaced by a fixed random unit vector of the same dimensionality (4096-d, matching Qwen3-Embedding-8B). For each training sample, each vector is generated deterministically by seeding a Gaussian sampler with a SHA-256 hash of the sample identifier and L2-normalizing the result; the same vector is reused without modification across all training epochs. The replacement occurs at the text encoder output: the random vectors are fed directly into the anchor projector $\pi_{an}$, which is trained normally. This setup tests whether the structural regularization induced by an auxiliary contrastive objective, independent of any linguistic content, is sufficient to explain the gains observed under semantic motion anchor supervision.

All three compared methods effectively perform text-to-text retrieval at inference, since the proxy is a description, but they differ in how the query and proxy are embedded. Text Contrastive has no anchor projector and routes both query and proxy through its transcript encoder into a single transcript-only space --- an easier setup that requires no cross-projector alignment. Our model and Random Anchor instead route the query through the transcript projector and the proxy through the anchor projector $\pi_{an}$, matching them in the learned shared space -- a stricter test of whether anchor supervision builds a transferable language--gesture representation. Random Anchor is itself a strong control: on BEAT2, it outperformed Text Contrastive (Table~\ref{tab:anchor}) by adding a structured though semantically uninformative auxiliary signal that regularized the shared space; carrying it forward here tests whether that regularization effect alone suffices for cross-domain transfer, or whether the semantic content of the anchors is what matters.

\subsection{Cross-Dataset Proxy Retrieval Metrics}
\label{app:cross_dataset_proxy_metrics}

Because TED and BEAT2 do not provide exact cross-dataset paired retrieval targets, we evaluate ranked results using two proxy relevance signals: a continuous semantic-context similarity score and a discrete exact semantic-label match. Let a query be denoted by $q$, and let the retrieved gallery items ranked by a model be $(g_1, g_2, \dots, g_K)$.

\paragraph{Semantic-context relevance.}
For each query $q$, we associate a semantic reference text $s(q)$ and for each gallery item $g$ we associate a gallery-side semantic reference $t(g)$. Both texts are embedded using a frozen Qwen3 text embedding model, yielding normalized vectors $\phi(s(q))$ and $\phi(t(g))$. We define the graded semantic relevance of gallery item $g$ to query $q$ as
\begin{equation}
r(q,g) = \max\!\left(0,\; \cos\!\big(\phi(s(q)), \phi(t(g))\big)\right).
\end{equation}
We clamp cosine values below zero to $0$ so that relevance is non-negative.

\paragraph{Cos@1.}
The top-1 semantic similarity is
\begin{equation}
\mathrm{Cos@1} = \frac{1}{|\mathcal{Q}|} \sum_{q \in \mathcal{Q}} r(q, g_1),
\end{equation}
where $\mathcal{Q}$ is the set of queries.

\paragraph{BestCos@K.}
For each query, we take the largest semantic relevance among the top-$K$ retrieved items:
\begin{equation}
\mathrm{BestCos@K} = \frac{1}{|\mathcal{Q}|} \sum_{q \in \mathcal{Q}} \max_{1 \le i \le K} r(q, g_i).
\end{equation}

\paragraph{MeanCos@K.}
For each query, we average semantic relevance over the top-$K$ retrieved items:
\begin{equation}
\mathrm{MeanCos@K} = \frac{1}{|\mathcal{Q}|} \sum_{q \in \mathcal{Q}} \frac{1}{K} \sum_{i=1}^{K} r(q, g_i).
\end{equation}

\paragraph{nDCG@K with semantic relevance.}
We use the semantic relevance values as graded gains. The discounted cumulative gain for query $q$ is
\begin{equation}
\mathrm{DCG@K}(q) = \sum_{i=1}^{K} \frac{r(q,g_i)}{\log_2(i+1)}.
\end{equation}
The ideal DCG is computed by sorting the full gallery for query $q$ by decreasing semantic relevance:
\begin{equation}
\mathrm{IDCG@K}(q) = \sum_{i=1}^{K} \frac{r^*_i(q)}{\log_2(i+1)},
\end{equation}
where $r^*_i(q)$ is the $i$-th largest relevance value available in the gallery for query $q$. We then compute
\begin{equation}
\mathrm{nDCG@K} = \frac{1}{|\mathcal{Q}|} \sum_{q \in \mathcal{Q}}
\frac{\mathrm{DCG@K}(q)}{\mathrm{IDCG@K}(q)}.
\end{equation}

\paragraph{Exact semantic-label relevance.}
For the exact-label evaluation, each query $q$ has a discrete semantic label $\ell(q)$ and each gallery item $g$ has a gallery label $\ell(g)$. We define binary relevance as
\begin{equation}
y(q,g) =
\begin{cases}
1, & \text{if } \ell(q)=\ell(g), \\
0, & \text{otherwise.}
\end{cases}
\end{equation}
These metrics are computed only on the shared TED/BEAT2 label space.

\paragraph{Acc@1.}
Top-1 exact-match accuracy is
\begin{equation}
\mathrm{Acc@1} = \frac{1}{|\mathcal{Q}|} \sum_{q \in \mathcal{Q}} y(q,g_1).
\end{equation}

\paragraph{Hit@K.}
A query counts as successful if at least one of the top-$K$ retrieved items has the correct label:
\begin{equation}
\mathrm{Hit@K} = \frac{1}{|\mathcal{Q}|} \sum_{q \in \mathcal{Q}}
\mathbb{I}\!\left[\max_{1 \le i \le K} y(q,g_i)=1\right].
\end{equation}

\paragraph{MRR.}
Let $\mathrm{rank}(q)$ be the rank of the first retrieved item whose label matches the query, and undefined if no match appears in the retrieved list. Then
\begin{equation}
\mathrm{MRR} = \frac{1}{|\mathcal{Q}|} \sum_{q \in \mathcal{Q}}
\begin{cases}
\frac{1}{\mathrm{rank}(q)}, & \text{if a match exists,} \\
0, & \text{otherwise.}
\end{cases}
\end{equation}

\paragraph{nDCG@K with semantic intent labels.}
For semantic intent-label evaluation, the gain at rank $i$ is binary:
\begin{equation}
\mathrm{DCG@K}(q) = \sum_{i=1}^{K} \frac{y(q,g_i)}{\log_2(i+1)}.
\end{equation}
The ideal ranking places all matching-label items first. If query $q$ has $M_q$ matching items in the gallery, then
\begin{equation}
\mathrm{IDCG@K}(q) = \sum_{i=1}^{\min(K,M_q)} \frac{1}{\log_2(i+1)}.
\end{equation}
The dataset-level metric is
\begin{equation}
\mathrm{nDCG@K} = \frac{1}{|\mathcal{Q}|} \sum_{q \in \mathcal{Q}}
\frac{\mathrm{DCG@K}(q)}{\mathrm{IDCG@K}(q)}.
\end{equation}

\paragraph{Pairwise win rate.}
For pairwise comparison between the proposed model and a baseline under metric $m$, we compute the metric separately for each query and define a win whenever the proposed score is larger than the baseline score. The win rate is
\begin{equation}
\mathrm{WinRate}(m) =
\frac{1}{|\mathcal{Q}|}
\sum_{q \in \mathcal{Q}}
\mathbb{I}\!\left[m_{\text{prop}}(q) > m_{\text{base}}(q)\right].
\end{equation}
Analogously, loss rate and tie rate are computed using $<$ and $=$ comparisons, respectively. In the main paper we report win rates primarily for the continuous semantic-context metrics, since binary exact-label metrics induce many ties and are therefore less informative in pairwise comparison.

\subsection{Additional Results: Rank distribution} 
\begin{minipage}[c]{0.44\linewidth}
\label{appx:cumrank}
Figure~\ref{fig:cumrank} plots the cumulative fraction of queries for which the ground-truth motion is retrieved among the top-$k$ candidates, across all rank cutoffs $k$. The proposed semantic-anchored model consistently yields a higher cumulative fraction than the direct text-motion baseline. Crucially, this advantage is heavily concentrated in the low-rank regime, precisely where a co-speech gesture system must commit to a single retrieval decision. While both models eventually recover most queries at higher ranks, the consistent gap at low $k$ indicates that semantic motion anchor supervision improves the overall ranking structure, not merely isolated operating points such as R@1.
\end{minipage}
\hfill
\begin{minipage}[c]{0.52\linewidth}
\centering
    \includegraphics[width=0.8\linewidth]{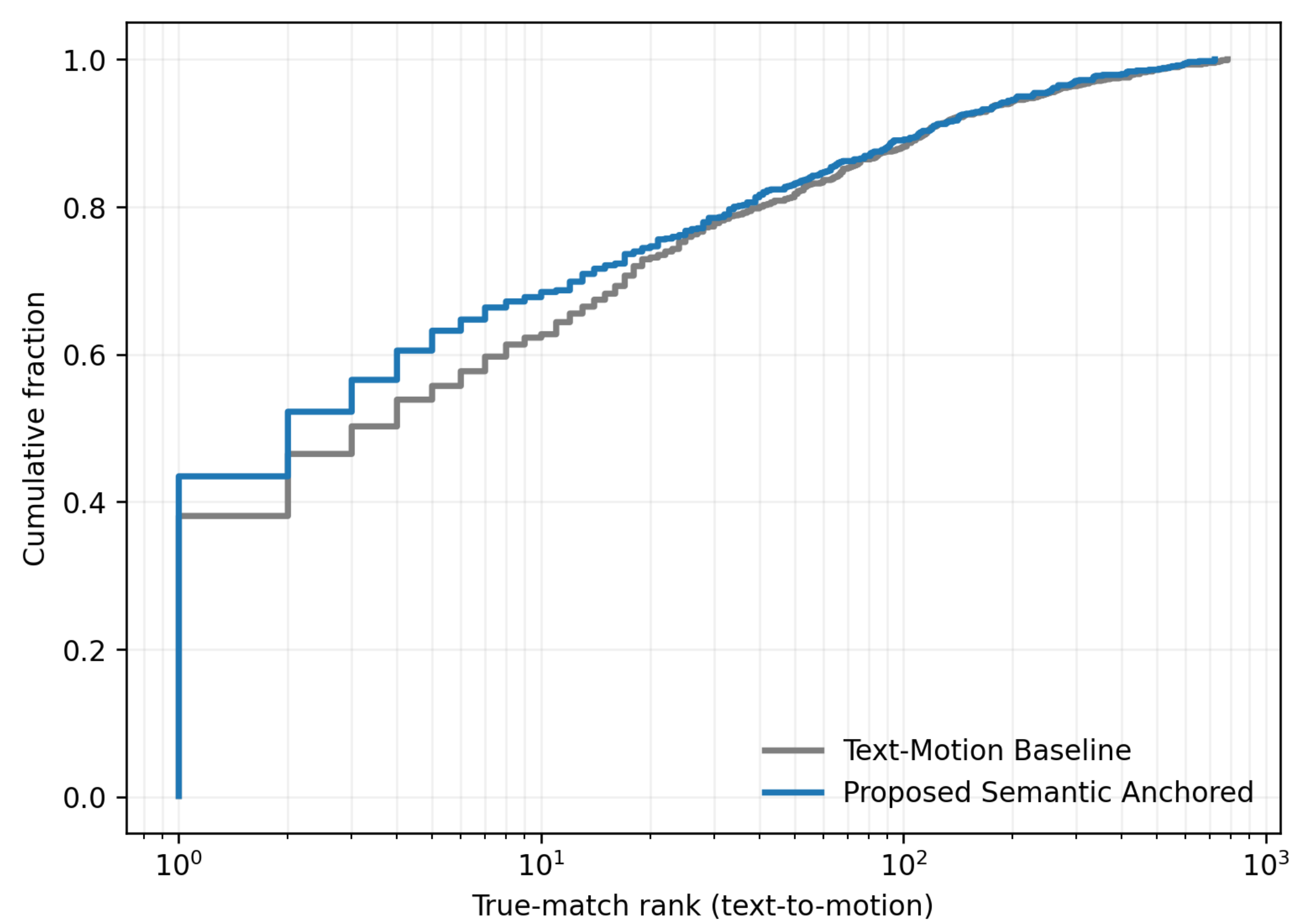}
    \captionof{figure}{
    Cumulative distribution of ground-truth ranks for text-to-motion retrieval. A higher curve indicates that more queries retrieve their paired ground-truth motion at lower ranks. The proposed semantic-anchored model consistently outperforms the text-motion baseline, with the largest gap in the low-rank regime, which dictates operational retrieval quality.
    }
    \label{fig:cumrank}
\end{minipage}

\subsection{Additional Results: Full Table of Effect of anchor content on retrieval performance} 
\label{appx:full-table-anchor-content}
Mismatched Anchors keeps the full objective and loss weights, but replaces $a^{phys}_i$ and $a^{int}_i$ with the anchors of two different training samples (never sample $i$ itself), using a fixed random assignment across all epochs. Physical Only uses $\mathcal{L}_{\text{phys}}$ as the only auxiliary loss ($\lambda_s{=}\lambda_b{=}0$), and Intent Only uses only $\mathcal{L}_{\text{int}}$ ($\lambda_p{=}\lambda_b{=}0$). All results are averaged over 7 seeds.
\FloatBarrier 
\begin{table*}[h]
\centering
\resizebox{\linewidth}{!}{%
\begin{tabular}{lccccccccc}
\toprule
 & \multicolumn{4}{c}{\textbf{Gesture $\rightarrow$ Text}} & \multicolumn{4}{c}{\textbf{Text $\rightarrow$ Gesture}} & \\
\cmidrule(lr){2-5} \cmidrule(lr){6-9}
\textbf{Method} & R@1 & R@5 & R@10 & MRR & R@1 & R@5 & R@10 & MRR & Avg. MRR [95\% CI] \\
\midrule
No Anchor & 37.2 & 57.5 & 65.4 & 47.0 & 39.1 & 58.7 & 66.3 & 48.5 & 47.73 [46.81, 48.65] \\
Random Anchor & $40.7$ & $60.3$ & $67.5$ & $50.0$ & $42.1$ & $61.0$ & $68.1$ & $51.2$ & 50.62 [50.16, 51.07]\\
Mismatched Anchors & 40.8 & 60.3 & 67.3 & 50.1 & 41.9 & 61.3 & 67.6 & 51.2 & 50.63 [50.02, 51.24] \\
Physical Only & 40.8 & 60.2 & 67.3 & 50.0 & 41.7 & 61.0 & 67.7 & 50.9 & 50.46 [49.75, 51.17] \\
Intent Only & $\mathbf{41.8}$ & $\mathbf{62.0}$ & $\mathbf{69.1}$ & $\mathbf{51.4}$ & $\mathbf{42.7}$ & $\mathbf{62.5}$ & 69.4 & $\mathbf{52.0}$ & $\mathbf{51.74}$ [51.13, 52.35] \\
\textbf{Ours} & $\mathbf{41.8}^{**}$ & $\mathbf{62.0}^{**}$ & $68.9^{*}$ & $\mathbf{51.4}^{\dagger}$ & $42.3$ & $\mathbf{62.5}^{**}$ & $\mathbf{69.5}^{**}$ & $51.9^{*}$ & 51.65 [51.03, 52.26] \\
\bottomrule
\multicolumn{10}{l}{\footnotesize $^{*}p<0.05$, $^{**}p<0.01$, $^{\dagger}p<0.001$}
\end{tabular}}
\caption{%
  Full anchor-content ablation on BEAT2. Significance markers on Ours indicate paired t-tests against Random Anchor: $^{*}p<0.05$, $^{**}p<0.01$, $^{\dagger}p<0.001$. Against Mismatched Anchors, Ours is also significantly better ($p<0.05$) on all metrics except Text$\rightarrow$Gesture R@1. Best displayed values are \textbf{bold}.%
  \label{tab:anchor2}
}
\end{table*}

Although Intent Only performs on par with Ours in-domain, only the physical branch aligns physical-form descriptions with the motion space. Unlike intent, these descriptions can be obtained from the motion alone, so they can serve as motion proxies in cross-dataset retrieval, where raw motion embeddings transfer poorly (Appendix~\ref{appx:cross8}).

\subsection{Additional Results: Cross-Dataset Retrieval in Both Directions}
\label{appx:cross8}
Section~\ref{sec:cross_dataset} uses TED transcripts as queries. Here, we evaluate the same BEAT2-trained models in both query--gallery directions between TED and BEAT2, with transcripts on one side and either raw motion embeddings ($z_m$) or physical-form proxies ($a^{phys}$) on the other, giving eight configurations (Table~\ref{tab:cross8}). Unlike Table~\ref{tab:ted_to_beat_emb_vs_proxy}, the proxy configurations represent BEAT2 gestures by $a^{phys}$ only, as for TED, and all results are averaged over 7 seeds with 95\% confidence intervals. With raw motion embeddings, all methods perform similarly (within 0.6 points on semantic nDCG@10). With physical-form proxies, our method ranks first on semantic nDCG@10 in three of the four configurations and is within 0.13 points of the best baseline in the fourth, which indicates that the physical-form proxy interface carries the cross-dataset advantage.

Compared with Intent Only, the full model also better aligns physical-form descriptions with raw motion. When $a^{phys}$ queries retrieve raw motion from the other dataset and relevance is measured by the similarity of physical-form descriptions (instead of semantic context in Appendix~\ref{app:cross_dataset_proxy_metrics}), nDCG@10 improves by $+1.25$ [$1.02$, $1.48$] for TED proxies to BEAT2 motion and $+1.21$ [$0.79$, $1.63$] for BEAT2 proxies to TED motion. It also retrieves better transcripts when physical-form proxies are used as queries. 
Together with Table~\ref{tab:anchor}, where Physical Only gives no in-domain gain over Random, this shows that neither single-branch variant (Intent Only or Physical Only) achieves both the in-domain gain and cross-dataset proxy transfer, whereas the full model with both branches does.

\FloatBarrier
\begin{table*}[h]
\centering

\resizebox{\textwidth}{!}{%
\begin{tabular}{l ccccc ccc}
\toprule
& \multicolumn{5}{c}{\textbf{Semantic Label (\%)}} & \multicolumn{3}{c}{\textbf{Semantic Context (\%)}} \\
\cmidrule(lr){2-6} \cmidrule(lr){7-9}
\textbf{Method} & Acc@1 & Hit@5 & Hit@10 & MRR & nDCG@10 & BestCos@5 & MeanCos@10 & nDCG@10 \\
\midrule
\multicolumn{9}{c}{\textbf{Raw motion embeddings ($z_m$)}} \\
\multicolumn{9}{l}{\textit{TED transcripts $\rightarrow$ BEAT2 $z_m$}} \\
Text Contrastive & 12.52{\tiny$\pm$1.42} & 35.75{\tiny$\pm$1.66} & 46.99{\tiny$\pm$1.43} & 23.64{\tiny$\pm$1.05} & 11.61{\tiny$\pm$0.46} & 55.87{\tiny$\pm$0.26} & 48.12{\tiny$\pm$0.14} & 69.52{\tiny$\pm$0.23} \\
Random Anchor & 11.92{\tiny$\pm$1.25} & 35.71{\tiny$\pm$1.59} & 47.41{\tiny$\pm$1.33} & 23.25{\tiny$\pm$1.09} & 11.74{\tiny$\pm$0.58} & 55.93{\tiny$\pm$0.28} & 48.12{\tiny$\pm$0.13} & 69.52{\tiny$\pm$0.21} \\
Intent Only & 12.39{\tiny$\pm$1.43} & 36.32{\tiny$\pm$1.64} & 47.65{\tiny$\pm$1.25} & 23.82{\tiny$\pm$1.08} & 12.05{\tiny$\pm$0.52} & 56.11{\tiny$\pm$0.27} & 48.31{\tiny$\pm$0.11} & 69.81{\tiny$\pm$0.23} \\
Proposed & 12.50{\tiny$\pm$1.51} & 36.56{\tiny$\pm$1.56} & 47.76{\tiny$\pm$1.64} & 23.85{\tiny$\pm$1.18} & 12.10$^{\dagger}${\tiny$\pm$0.57} & 56.11$^{\dagger}${\tiny$\pm$0.24} & 48.30$^{\dagger}${\tiny$\pm$0.10} & 69.79$^{\dagger}${\tiny$\pm$0.19} \\
\midrule
\multicolumn{9}{l}{\textit{BEAT2 transcripts $\rightarrow$ TED $z_m$}} \\
Text Contrastive & 9.23{\tiny$\pm$1.08} & 38.20{\tiny$\pm$2.48} & 61.25{\tiny$\pm$2.76} & 23.88{\tiny$\pm$1.35} & 9.74{\tiny$\pm$0.48} & 54.36{\tiny$\pm$0.27} & 47.24{\tiny$\pm$0.13} & 74.91{\tiny$\pm$0.27} \\
Random Anchor & 9.75{\tiny$\pm$0.62} & 39.15{\tiny$\pm$2.11} & 62.13{\tiny$\pm$2.07} & 24.41{\tiny$\pm$0.81} & 9.93{\tiny$\pm$0.28} & 54.42{\tiny$\pm$0.17} & 47.26{\tiny$\pm$0.13} & 74.93{\tiny$\pm$0.24} \\
Intent Only & 9.21{\tiny$\pm$1.01} & 38.82{\tiny$\pm$1.71} & 61.67{\tiny$\pm$2.14} & 24.22{\tiny$\pm$1.15} & 9.90{\tiny$\pm$0.35} & 54.40{\tiny$\pm$0.14} & 47.25{\tiny$\pm$0.13} & 74.93{\tiny$\pm$0.27} \\
Proposed & 9.36{\tiny$\pm$0.90} & 38.53{\tiny$\pm$1.87} & 61.35{\tiny$\pm$3.16} & 24.15{\tiny$\pm$1.10} & 9.89{\tiny$\pm$0.43} & 54.39{\tiny$\pm$0.19} & 47.27{\tiny$\pm$0.15} & 74.94{\tiny$\pm$0.27} \\
\midrule
\multicolumn{9}{l}{\textit{BEAT2 $z_m$ $\rightarrow$ TED transcripts}} \\
Text Contrastive & 12.05{\tiny$\pm$1.02} & 41.27{\tiny$\pm$1.19} & 61.08{\tiny$\pm$1.06} & 26.54{\tiny$\pm$0.96} & 11.23{\tiny$\pm$0.48} & 54.73{\tiny$\pm$0.17} & 47.64{\tiny$\pm$0.22} & 75.59{\tiny$\pm$0.43} \\
Random Anchor & 11.92{\tiny$\pm$1.06} & 41.76{\tiny$\pm$1.36} & 61.52{\tiny$\pm$2.29} & 26.41{\tiny$\pm$1.05} & 11.26{\tiny$\pm$0.48} & 54.67{\tiny$\pm$0.30} & 47.64{\tiny$\pm$0.24} & 75.58{\tiny$\pm$0.46} \\
Intent Only & 11.60{\tiny$\pm$1.24} & 41.41{\tiny$\pm$1.54} & 61.50{\tiny$\pm$0.83} & 26.32{\tiny$\pm$1.06} & 11.35{\tiny$\pm$0.47} & 54.83{\tiny$\pm$0.25} & 47.70{\tiny$\pm$0.24} & 75.67{\tiny$\pm$0.47} \\
Proposed & 11.68{\tiny$\pm$1.32} & 41.20{\tiny$\pm$1.61} & 61.62{\tiny$\pm$1.62} & 26.27{\tiny$\pm$1.13} & 11.38{\tiny$\pm$0.52} & 54.77{\tiny$\pm$0.23} & 47.68{\tiny$\pm$0.22} & 75.63{\tiny$\pm$0.45} \\
\midrule
\multicolumn{9}{l}{\textit{TED $z_m$ $\rightarrow$ BEAT2 transcripts}} \\
Text Contrastive & 10.08{\tiny$\pm$1.70} & 34.04{\tiny$\pm$1.05} & 46.62{\tiny$\pm$1.52} & 21.58{\tiny$\pm$1.12} & 10.03{\tiny$\pm$0.62} & 56.39{\tiny$\pm$0.67} & 48.34{\tiny$\pm$0.38} & 70.00{\tiny$\pm$0.71} \\
Random Anchor & 9.99{\tiny$\pm$1.83} & 33.60{\tiny$\pm$1.59} & 45.26{\tiny$\pm$1.22} & 21.23{\tiny$\pm$1.39} & 10.01{\tiny$\pm$0.62} & 55.78{\tiny$\pm$0.69} & 48.02{\tiny$\pm$0.51} & 69.45{\tiny$\pm$0.73} \\
Intent Only & 10.03{\tiny$\pm$1.54} & 34.30{\tiny$\pm$1.35} & 46.53{\tiny$\pm$1.59} & 21.56{\tiny$\pm$1.06} & 10.27{\tiny$\pm$0.54} & 55.99{\tiny$\pm$0.51} & 48.23{\tiny$\pm$0.54} & 69.76{\tiny$\pm$0.66} \\
Proposed & 9.70{\tiny$\pm$1.47} & 33.55{\tiny$\pm$1.14} & 45.50{\tiny$\pm$1.39} & 21.16{\tiny$\pm$1.01} & 9.97{\tiny$\pm$0.69} & 55.99{\tiny$\pm$0.52} & 48.14{\tiny$\pm$0.62} & 69.65{\tiny$\pm$0.84} \\
\midrule[\heavyrulewidth]
\multicolumn{9}{c}{\textbf{Physical-form proxies ($a^{phys}$)}} \\
\multicolumn{9}{l}{\textit{TED transcripts $\rightarrow$ BEAT2 $a^{phys}$}} \\
Text Contrastive & 11.48{\tiny$\pm$1.66} & 34.10{\tiny$\pm$2.05} & 46.58{\tiny$\pm$1.61} & 22.52{\tiny$\pm$1.50} & 10.54{\tiny$\pm$0.73} & 56.58{\tiny$\pm$1.03} & 48.01{\tiny$\pm$0.61} & 69.36{\tiny$\pm$0.98} \\
Random Anchor & 10.71{\tiny$\pm$1.00} & 34.17{\tiny$\pm$1.18} & 45.28{\tiny$\pm$1.00} & 21.99{\tiny$\pm$0.89} & 10.67{\tiny$\pm$0.52} & 55.90{\tiny$\pm$0.84} & 47.39{\tiny$\pm$0.55} & 68.40{\tiny$\pm$0.86} \\
Intent Only & 11.42{\tiny$\pm$0.73} & 36.43{\tiny$\pm$1.76} & 47.87{\tiny$\pm$1.81} & 23.22{\tiny$\pm$0.82} & 11.57{\tiny$\pm$0.49} & 57.52{\tiny$\pm$0.71} & 48.77{\tiny$\pm$0.38} & 70.29{\tiny$\pm$0.54} \\
Proposed & \textbf{11.86}{\tiny$\pm$1.01} & \textbf{37.00}$^{\dagger}${\tiny$\pm$1.94} & \textbf{49.06}$^{\dagger}${\tiny$\pm$1.31} & \textbf{23.67}{\tiny$\pm$0.80} & \textbf{11.75}$^{\dagger}${\tiny$\pm$0.45} & \textbf{57.97}$^{\dagger}${\tiny$\pm$0.21} & \textbf{48.96}$^{\dagger}${\tiny$\pm$0.18} & \textbf{70.68}$^{\dagger}${\tiny$\pm$0.26} \\
\midrule
\multicolumn{9}{l}{\textit{BEAT2 transcripts $\rightarrow$ TED $a^{phys}$}} \\
Text Contrastive & 9.35{\tiny$\pm$1.25} & 31.59{\tiny$\pm$1.99} & 48.88{\tiny$\pm$1.90} & 21.32{\tiny$\pm$1.41} & \textbf{10.54}{\tiny$\pm$0.88} & 53.71{\tiny$\pm$0.28} & 47.28{\tiny$\pm$0.30} & 74.84{\tiny$\pm$0.54} \\
Random Anchor & 8.31{\tiny$\pm$0.74} & 28.05{\tiny$\pm$1.34} & 45.24{\tiny$\pm$1.50} & 19.41{\tiny$\pm$0.84} & 8.89{\tiny$\pm$0.60} & 53.01{\tiny$\pm$0.19} & 46.83{\tiny$\pm$0.23} & 74.10{\tiny$\pm$0.36} \\
Intent Only & \textbf{9.86}{\tiny$\pm$1.44} & \textbf{33.24}{\tiny$\pm$1.15} & \textbf{51.59}{\tiny$\pm$0.59} & \textbf{22.26}{\tiny$\pm$1.03} & 10.36{\tiny$\pm$0.38} & 54.40{\tiny$\pm$0.25} & 47.62{\tiny$\pm$0.21} & 75.57{\tiny$\pm$0.36} \\
Proposed & 8.91{\tiny$\pm$0.91} & 31.46{\tiny$\pm$1.66} & 49.03{\tiny$\pm$1.99} & 21.04{\tiny$\pm$0.99} & 10.27{\tiny$\pm$0.62} & \textbf{54.44}$^{\dagger}${\tiny$\pm$0.16} & \textbf{47.67}$^{\dagger}${\tiny$\pm$0.13} & \textbf{75.60}$^{\dagger}${\tiny$\pm$0.24} \\
\midrule
\multicolumn{9}{l}{\textit{BEAT2 $a^{phys}$ $\rightarrow$ TED transcripts}} \\
Text Contrastive & 9.53{\tiny$\pm$4.75} & 36.23{\tiny$\pm$5.74} & 55.99{\tiny$\pm$5.93} & 23.26{\tiny$\pm$5.02} & 9.47{\tiny$\pm$0.99} & 54.60{\tiny$\pm$0.76} & \textbf{47.79}{\tiny$\pm$0.69} & \textbf{75.89}{\tiny$\pm$1.27} \\
Random Anchor & \textbf{11.53}{\tiny$\pm$6.04} & \textbf{42.39}{\tiny$\pm$8.33} & \textbf{63.87}{\tiny$\pm$7.21} & \textbf{26.58}{\tiny$\pm$5.94} & \textbf{11.35}{\tiny$\pm$2.78} & 54.33{\tiny$\pm$0.98} & 47.27{\tiny$\pm$1.31} & 74.90{\tiny$\pm$2.46} \\
Intent Only & 5.74{\tiny$\pm$1.62} & 29.47{\tiny$\pm$7.77} & 48.88{\tiny$\pm$9.82} & 18.38{\tiny$\pm$2.39} & 8.35{\tiny$\pm$1.13} & 53.03{\tiny$\pm$0.64} & 46.21{\tiny$\pm$0.65} & 72.69{\tiny$\pm$1.58} \\
Proposed & 10.03{\tiny$\pm$0.85} & 40.20{\tiny$\pm$2.53} & 60.63{\tiny$\pm$2.69} & 24.85{\tiny$\pm$1.07} & 11.08{\tiny$\pm$0.59} & \textbf{54.95}{\tiny$\pm$0.33} & 47.78{\tiny$\pm$0.35} & 75.76{\tiny$\pm$0.64} \\
\midrule
\multicolumn{9}{l}{\textit{TED $a^{phys}$ $\rightarrow$ BEAT2 transcripts}} \\
Text Contrastive & 10.78{\tiny$\pm$1.96} & 36.82{\tiny$\pm$1.89} & 49.74{\tiny$\pm$2.06} & 22.85{\tiny$\pm$1.14} & 11.00{\tiny$\pm$0.54} & 56.84{\tiny$\pm$1.25} & 48.39{\tiny$\pm$0.81} & 70.03{\tiny$\pm$1.43} \\
Random Anchor & 11.22{\tiny$\pm$2.59} & 34.74{\tiny$\pm$3.31} & 47.23{\tiny$\pm$3.91} & 22.58{\tiny$\pm$2.07} & 10.73{\tiny$\pm$0.98} & 55.71{\tiny$\pm$1.47} & 47.59{\tiny$\pm$0.98} & 68.74{\tiny$\pm$1.65} \\
Intent Only & 11.20{\tiny$\pm$2.38} & 38.01{\tiny$\pm$3.28} & 49.85{\tiny$\pm$2.88} & 23.42{\tiny$\pm$2.43} & 11.39{\tiny$\pm$0.78} & 55.51{\tiny$\pm$0.74} & 47.91{\tiny$\pm$0.56} & 69.30{\tiny$\pm$0.92} \\
Proposed & \textbf{12.08}{\tiny$\pm$1.94} & \textbf{41.00}$^{\dagger}${\tiny$\pm$1.95} & \textbf{54.57}$^{\dagger}${\tiny$\pm$1.47} & \textbf{25.28}{\tiny$\pm$1.61} & \textbf{12.82}$^{\dagger}${\tiny$\pm$0.56} & \textbf{57.08}{\tiny$\pm$0.40} & \textbf{49.03}{\tiny$\pm$0.19} & \textbf{70.98}{\tiny$\pm$0.30} \\
\bottomrule
\end{tabular}}
\caption{Cross-dataset retrieval in all eight query--gallery configurations between TED and BEAT2 (mean {\tiny$\pm$} half-width of the 95\% CI over 7 seeds), grouped by whether motion is represented by raw motion embeddings ($z_m$) or physical-form proxies ($a^{phys}$); each block is labeled query $\rightarrow$ gallery. All models are trained only on BEAT2. Semantic label and semantic-context metrics follow Appendix~\ref{app:cross_dataset_proxy_metrics}. Best results in the proxy configurations are \textbf{bold}. $^{\dagger}$Proposed is significantly better than both Text Contrastive and Random Anchor (paired t-test, $p<0.05$).}
\label{tab:cross8}
\end{table*}

\subsection{Additional Results: Semantic-Context Metrics with an Independent Encoder}
\label{appx:bge_m3}
The semantic-context metrics in Table~\ref{tab:ted_to_beat_emb_vs_proxy} measure relevance with Qwen3 embeddings, the same model family as the frozen text encoder of the retrieval models. To rule out a bias toward this embedding space, we keep the retrieval rankings of every model fixed and recompute relevance with BGE-M3, an independent embedding model (Table~\ref{tab:bge_m3}). Semantic-label metrics do not depend on the embedding model and are unchanged. Our method remains first on all three semantic-context metrics in both gallery settings.

\FloatBarrier
\begin{table}[h]
\centering
\small
\begin{tabular}{lccc}
\toprule
\textbf{Method} & BestCos@5 & MeanCos@10 & nDCG@10 \\
\midrule
\multicolumn{4}{l}{\textit{TED transcripts $\rightarrow$ BEAT2 motion embeddings ($z_m$)}} \\
Text Contrastive & 58.22 & 50.84 & 72.87 \\
Random Anchor & 58.21 & 50.94 & 72.97 \\
Mismatched Anchors & 58.14 & 51.00 & 73.02 \\
Proposed & \textbf{58.41} & \textbf{51.05} & \textbf{73.16} \\
\midrule
\multicolumn{4}{l}{\textit{TED transcripts $\rightarrow$ BEAT2 semantic anchor proxies}} \\
Text Contrastive & 58.29 & 51.32 & 73.43 \\
Random Anchor & 55.74 & 49.52 & 70.86 \\
Mismatched Anchors & 56.53 & 50.02 & 71.51 \\
Proposed & \textbf{58.66} & \textbf{51.91} & \textbf{74.21} \\
\bottomrule
\end{tabular}
\caption{TED-to-BEAT2 semantic-context metrics (\%) with relevance recomputed by BGE-M3, using the same retrieval rankings as Table~\ref{tab:ted_to_beat_emb_vs_proxy}. Best results per gallery setting are \textbf{bold}.}
\label{tab:bge_m3}
\end{table}

\subsection{Additional Results: Full Table of TED-to-TED}
\label{appx:full-table-ted-transfer}
\FloatBarrier 
\begin{table*}[h]
\centering
\resizebox{\textwidth}{!}{%
\begin{tabular}{llcccccccc}
\toprule
 & & \multicolumn{4}{c}{\textbf{Gesture $\rightarrow$ Text}} & \multicolumn{4}{c}{\textbf{Text $\rightarrow$ Gesture}} \\
\cmidrule(lr){3-6} \cmidrule(lr){7-10}
\textbf{Method} & \textbf{Gesture Rep.} & R@1 & R@5 & R@10 & MRR & R@1 & R@5 & R@10 & MRR \\
\midrule
Text Contrastive & Motion Embed. ($z_m$) & 0.3 & 1.0 & 1.5 & 1.10 & 0.0 & 0.5 & 1.2 & 0.82 \\
Random Anchor & Motion Embed. ($z_m$) & 0.1 & 1.3 & 1.8 & 1.17 & 0.1 & 0.4 & 1.5 & 0.96 \\
Proposed Semantic Anchored & Motion Embed. ($z_m$) & 0.1 & 1.0 & 1.5 & 1.12 & 0.0 & 0.5 & 1.9 & 0.91 \\
\midrule
Text Contrastive $\dagger$ & Motion description ($a^{phys}$) & 0.6 & 1.8 & 3.0 & 1.91 & 1.0 & 2.8 & 5.0 & 2.64 \\
Random Anchor & Motion description ($a^{phys}$) & 0.1 & 0.5 & 0.6 & 0.80 & 0.1 & 0.5 & 0.5 & 0.79 \\
Proposed Semantic Anchored & Motion description ($a^{phys}$) & \textbf{1.2} & \textbf{4.6} & \textbf{7.1} & \textbf{3.48} & \textbf{1.2} & \textbf{4.5} & \textbf{6.0} & \textbf{3.42} \\

\bottomrule
\end{tabular}%
}
\caption{
Proxy-based cross-dataset generalization on TED. Comparison when the gallery gestures are represented by the physical-form motion anchor ($a^{phys}$) compared to motion embeddings. Best results are \textbf{bold}. $\dagger$ Here, the motion description is passed via the transcript encoder.
}
\label{appx:table_ted_generalisation_proxies}
\end{table*}

\subsection{Motion Distribution Shift between BEAT2 and TED}
\label{appx:domain_shift}
BEAT2 and TED motions are both expressed as SMPL-X pose parameters, but they come from different acquisition and processing pipelines: studio motion capture for BEAT2, and monocular video pose estimation, segmentation and preprocessing for TED. We measure how different the resulting motion distributions are, and whether these differences explain the near-chance transfer of the raw motion encoder in Section~\ref{sec:cross_dataset}.

\paragraph{Domain probe.} A logistic-regression probe on per-clip pose statistics, evaluated out-of-fold with BEAT2 speakers and TED videos as groups, separates BEAT2 from TED with ROC AUC 0.999 (95\% group-bootstrap CI [0.998, 1.000]). This holds even after removing sequence length and root orientation, resampling every clip to 32 frames, mean-centering each clip over time, normalizing its motion scale, and matching semantic-label frequencies between the two datasets.

\paragraph{Input statistics.} TED clips are much shorter than BEAT2 clips (median 20 vs.\ 330 frames). Relative to BEAT2, TED also has 0.46$\times$ the hand-pose RMS magnitude, 0.33$\times$ the within-clip dynamic RMS and 2.61$\times$ the per-frame acceleration RMS (medians over clips).

\paragraph{Perturbation test.} To test whether these differences affect the raw motion encoder, we perturb BEAT2 test motion to match each TED statistic and evaluate the trained model without retraining. Shortening clips to 20 frames reduces average MRR by 36.6 points, attenuating the hand components by 22.2 points and halving the temporal sampling by 20.0 points; the combined TED-matched transformation reduces it by 43.1 points, whereas small pose jitter causes no measurable degradation. Sharing the SMPL-X parameterization therefore does not make the two motion distributions comparable, and the raw motion encoder is sensitive to precisely the properties in which they differ. Closing this gap would require motion harmonization or cross-estimator domain adaptation, which is outside the scope of this work.


\subsection{Details on Perceptual User Study}
\label{appx:userstudy}
%
%
\begin{minipage}[h]{0.44\linewidth}
We conducted two user studies to evaluate our approach. In the first, 32 anonymous participants, primarily comprising university academic staff and students, were presented with 10 forced-choice questions using an online evaluation form. Each question displayed a side-by-side gestural animation comparing gestures retrieved by our method against those retrieved by RAG-Gesture’s LLM-based gesture type retrieval approach. We utilized the same input word query as \cite{mughal2025retrieving} and replaced their word-to-motion retrieval framework with our anchor-based retrieval for the given word. For each pair, participants answered: \textit{"Which of the two gestures better suits the red highlighted word written above?"}

In the second study, 16 anonymous participants evaluated end-to-end gesture generation quality using the same forced-choice format. Retrieved gestures from both methods were passed through an identical generator; only the retrieval step differed. For each pair, participants answered two questions: (a) \textit{"Which of the two gestures better suits the red highlighted word written above?"} to assess semantic appropriateness, and (b) \textit{"Which of the two gestures looks more natural?"} to assess motion quality. The query word was highlighted in red above each animation pair, as shown in Figure~\ref{userstudy:screenshot}.

\end{minipage}
\hfill
\begin{minipage}[h]{0.52\linewidth}
\centering
    \includegraphics[width=\linewidth]{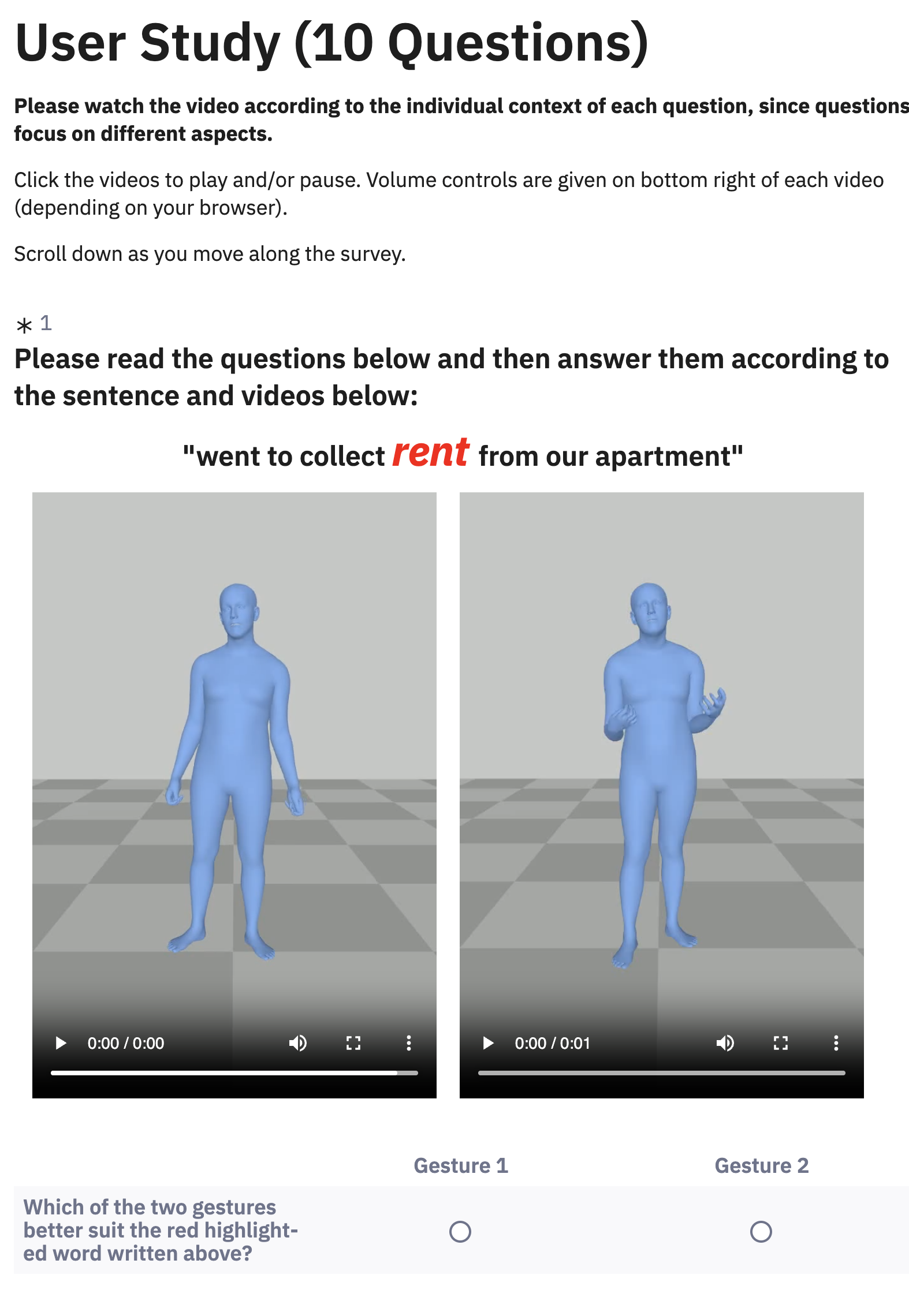}
    \captionof{figure}{
    \textbf{Screen Capture of User Study Interface.} The query input word is highlighted in red.
    }
    \label{userstudy:screenshot}
\end{minipage}

\newpage
\begin{table*}[h!]
\centering
\small
\begin{tabular}{p{\linewidth}}
\hline
\textbf{Example 1 --- \textit{Self-reference}} \\
"\ldots I think \textbf{my favourite thing} about Halloween is the haunted houses, I'm a big fan of the rush of feeling scared\ldots \\
\textbf{GT:} One hand rises to chest level and settles near the \ldots, marking a personal emphasis. It conveys the speaker's own enthusiasm and fondness for the thrill of haunted houses. \\
\textbf{Ours:} Both hands lift slightly from below the waist and hover \ldots. This conveys heightened positive emotion, emphasizing how happy and excited the speaker felt. \\
\textbf{Text \& Random:} Both hands rise to chest level and spread outward, then settle lower and apart with open palms, as if laying out a broad area. This depicts something being widely present \ldots just one time period. \\

\hline

\textbf{Example 2 --- \textit{Other-reference}} \\
"\ldots I went \textbf{back to the hotel} that we were staying at, I'm back to where I last saw them\ldots" \\
\textbf{GT:} One hand extends outward and then lowers \ldots , as if marking a location in space, to indicate returning to the place where the speaker last saw them. \\
\textbf{Ours:} One hand points outward at chest level \ldots a specific place or person being referred to in the story. \\
\textbf{Text \& Random:} Both hands \ldots, as if presenting the situation in front of the speaker. The gesture conveys an offering or reference to the bad circumstances happening to other people. \\

\hline

\textbf{Example 3 --- \textit{Other-reference}} \\
"\ldots and then \textbf{a girl} I think she was in like university or something\ldots" \\
\textbf{GT:} One hand is \ldots in a small presenting gesture, as if introducing or referring to the girl being mentioned. \\
\textbf{Ours:} One hand extends slightly forward \ldots, as if indicating a person positioned ahead. It serves to point out the girl sitting in front within the described scene. \\
\textbf{Text:} Both hands spread slightly outward \ldots, palm-up presentation, conveying the idea of a limited number of social festivals and presenting this as a general observation. \\
\textbf{Random:} One hand lifts and makes a small \ldots, as if marking off another item. It serves to single out "tennis" as an additional example in the list. \\

\hline

\textbf{Example 4 --- \textit{Symbolise}} \\
"\ldots it uses a very \textbf{long weapon}, like a big big big sword\ldots" \\
\textbf{GT:} One hand sweeps outward and rises \ldots level, briefly pointing as it extends to depict the great length and imposing size of the imagined sword. \\
\textbf{Ours:} Both hands are held apart in front of the body, \ldots, with open palms, to convey the largeness or broad significance of what is being described. \\
\textbf{Text \& Random:} Both hands \ldots turned out, marking a contrast. The gesture helps separate and dismiss "the job" itself as not being the source of the problem. \\

\hline

\textbf{Example 5 --- \textit{Symbolise}} \\
"\ldots it's amazing because it's \textbf{huge}, I don't know how many people have the day off at the same time\ldots" \\
\textbf{GT:} Both hands sweep upward and outward \ldots and lower outward with open palms, to depict the great size and expansiveness of the place. \\
\textbf{Ours:} Both hands move outward and down from \ldots, then hold wide apart. The gesture presents a broad set of possibilities or "all these answers," emphasizing their collective scope. \\
\textbf{Text \& Random:} Both \ldots, with open palms facing outward, to negate or dismiss the idea of a separate "real life."\\

\hline

\textbf{Example 6 --- \textit{Enumeration}} \\
"\ldots I'll actually \textbf{give each restaurant a score} based on how good the food is\ldots" \\
\textbf{GT:} Both hands rise briefly to chest level \ldots slightly more active as if marking off items. The gesture conveys the speaker's intention to assign or enumerate a score for each restaurant. \\
\textbf{Ours:} Both hands lower and hold forward \ldots, as if laying out a set of rules. This presents the category of acceptable versus unacceptable behavior and underscores clear boundaries. \\
\textbf{Text:} One hand lifts \ldots slightly toward the speaker, while the other remains low and still, to refer to spending time with oneself in a personal, inviting way. \textbf{[Self-reference]} \\
\textbf{Random:} One hand lifts and moves \ldots, as if casually indicating a usual destination. It conveys a matter-of-fact reference to the speaker's routine of going to the library. \\
\hline
\textbf{Example 7 --- \textit{Symbolise}} \\
"\ldots in the middle of my room there is a \textbf{big soft bed}\ldots" \\
\textbf{GT:} Both hands rise \ldots in a broad, relaxed spread as if laying out space. The gesture depicts the bed's large, expansive presence in the room. \\
\textbf{Ours:} Both hands \ldots and spread down toward the waist. The gesture depicts laying out or displaying the pictures on the wall. \\
\textbf{Text:} Both hands \ldots, with one hand shifting palm-up to the side, to indicate the other side of the room and present the small desk located there. \\
\textbf{Random:} Both hands move down and slightly \ldots, held open as if presenting a spread of items. This depicts the accumulation or abundance of clothes, bags, and shoes. \\
\bottomrule
\end{tabular}
\caption{Additional Qualitative Retrieval Examples}
\label{tab:qual_eg_appendix}
\end{table*}






\newpage
\section*{NeurIPS Paper Checklist}

\begin{enumerate}

\item {\bf Claims}
    \item[] Question: Do the main claims made in the abstract and introduction accurately reflect the paper's contributions and scope?
    \item[] Answer: \answerYes{} 
    \item[] Justification: We have clearly stated the claims and contributions in the abstract and introduction.
    \item[] Guidelines:
    \begin{itemize}
        \item The answer \answerNA{} means that the abstract and introduction do not include the claims made in the paper.
        \item The abstract and/or introduction should clearly state the claims made, including the contributions made in the paper and important assumptions and limitations. A \answerNo{} or \answerNA{} answer to this question will not be perceived well by the reviewers. 
        \item The claims made should match theoretical and experimental results, and reflect how much the results can be expected to generalize to other settings. 
        \item It is fine to include aspirational goals as motivation as long as it is clear that these goals are not attained by the paper. 
    \end{itemize}

\item {\bf Limitations}
    \item[] Question: Does the paper discuss the limitations of the work performed by the authors?
    \item[] Answer: \answerYes{} 
    \item[] Justification: We have discussed in the Limitation Section~\ref{sec:limitation}
    \item[] Guidelines:
    \begin{itemize}
        \item The answer \answerNA{} means that the paper has no limitation while the answer \answerNo{} means that the paper has limitations, but those are not discussed in the paper. 
        \item The authors are encouraged to create a separate ``Limitations'' section in their paper.
        \item The paper should point out any strong assumptions and how robust the results are to violations of these assumptions (e.g., independence assumptions, noiseless settings, model well-specification, asymptotic approximations only holding locally). The authors should reflect on how these assumptions might be violated in practice and what the implications would be.
        \item The authors should reflect on the scope of the claims made, e.g., if the approach was only tested on a few datasets or with a few runs. In general, empirical results often depend on implicit assumptions, which should be articulated.
        \item The authors should reflect on the factors that influence the performance of the approach. For example, a facial recognition algorithm may perform poorly when image resolution is low or images are taken in low lighting. Or a speech-to-text system might not be used reliably to provide closed captions for online lectures because it fails to handle technical jargon.
        \item The authors should discuss the computational efficiency of the proposed algorithms and how they scale with dataset size.
        \item If applicable, the authors should discuss possible limitations of their approach to address problems of privacy and fairness.
        \item While the authors might fear that complete honesty about limitations might be used by reviewers as grounds for rejection, a worse outcome might be that reviewers discover limitations that aren't acknowledged in the paper. The authors should use their best judgment and recognize that individual actions in favor of transparency play an important role in developing norms that preserve the integrity of the community. Reviewers will be specifically instructed to not penalize honesty concerning limitations.
    \end{itemize}

\item {\bf Theory assumptions and proofs}
    \item[] Question: For each theoretical result, does the paper provide the full set of assumptions and a complete (and correct) proof?
    \item[] Answer: \answerNA{} 
    \item[] Justification: The paper does not have theoretical results.
    \item[] Guidelines:
    \begin{itemize}
        \item The answer \answerNA{} means that the paper does not include theoretical results. 
        \item All the theorems, formulas, and proofs in the paper should be numbered and cross-referenced.
        \item All assumptions should be clearly stated or referenced in the statement of any theorems.
        \item The proofs can either appear in the main paper or the supplemental material, but if they appear in the supplemental material, the authors are encouraged to provide a short proof sketch to provide intuition. 
        \item Inversely, any informal proof provided in the core of the paper should be complemented by formal proofs provided in appendix or supplemental material.
        \item Theorems and Lemmas that the proof relies upon should be properly referenced. 
    \end{itemize}

    \item {\bf Experimental result reproducibility}
    \item[] Question: Does the paper fully disclose all the information needed to reproduce the main experimental results of the paper to the extent that it affects the main claims and/or conclusions of the paper (regardless of whether the code and data are provided or not)?
    \item[] Answer: \answerYes{} 
    \item[] Justification: The paper includes all necessary details for reproducing the main experimental results in section~\ref{sec:evaluation} and Appendix.
    \item[] Guidelines:
    \begin{itemize}
        \item The answer \answerNA{} means that the paper does not include experiments.
        \item If the paper includes experiments, a \answerNo{} answer to this question will not be perceived well by the reviewers: Making the paper reproducible is important, regardless of whether the code and data are provided or not.
        \item If the contribution is a dataset and\slash or model, the authors should describe the steps taken to make their results reproducible or verifiable. 
        \item Depending on the contribution, reproducibility can be accomplished in various ways. For example, if the contribution is a novel architecture, describing the architecture fully might suffice, or if the contribution is a specific model and empirical evaluation, it may be necessary to either make it possible for others to replicate the model with the same dataset, or provide access to the model. In general. releasing code and data is often one good way to accomplish this, but reproducibility can also be provided via detailed instructions for how to replicate the results, access to a hosted model (e.g., in the case of a large language model), releasing of a model checkpoint, or other means that are appropriate to the research performed.
        \item While NeurIPS does not require releasing code, the conference does require all submissions to provide some reasonable avenue for reproducibility, which may depend on the nature of the contribution. For example
        \begin{enumerate}
            \item If the contribution is primarily a new algorithm, the paper should make it clear how to reproduce that algorithm.
            \item If the contribution is primarily a new model architecture, the paper should describe the architecture clearly and fully.
            \item If the contribution is a new model (e.g., a large language model), then there should either be a way to access this model for reproducing the results or a way to reproduce the model (e.g., with an open-source dataset or instructions for how to construct the dataset).
            \item We recognize that reproducibility may be tricky in some cases, in which case authors are welcome to describe the particular way they provide for reproducibility. In the case of closed-source models, it may be that access to the model is limited in some way (e.g., to registered users), but it should be possible for other researchers to have some path to reproducing or verifying the results.
        \end{enumerate}
    \end{itemize}

\item {\bf Open access to data and code}
    \item[] Question: Does the paper provide open access to the data and code, with sufficient instructions to faithfully reproduce the main experimental results, as described in supplemental material?
    \item[] Answer: \answerYes{} 
    \item[] Justification: The code, experiment configurations and instructions to reproduce the main results are released at \url{https://github.com/aboots/Semantic_Motion_Anchors}, and the \textsc{Semantix} dataset together with the generated semantic motion anchors is released at \url{https://huggingface.co/datasets/aboots/SEMANTIX}; both are linked from the project page.
    \item[] Guidelines:
    \begin{itemize}
        \item The answer \answerNA{} means that paper does not include experiments requiring code.
        \item Please see the NeurIPS code and data submission guidelines (\url{https://neurips.cc/public/guides/CodeSubmissionPolicy}) for more details.
        \item While we encourage the release of code and data, we understand that this might not be possible, so \answerNo{} is an acceptable answer. Papers cannot be rejected simply for not including code, unless this is central to the contribution (e.g., for a new open-source benchmark).
        \item The instructions should contain the exact command and environment needed to run to reproduce the results. See the NeurIPS code and data submission guidelines (\url{https://neurips.cc/public/guides/CodeSubmissionPolicy}) for more details.
        \item The authors should provide instructions on data access and preparation, including how to access the raw data, preprocessed data, intermediate data, and generated data, etc.
        \item The authors should provide scripts to reproduce all experimental results for the new proposed method and baselines. If only a subset of experiments are reproducible, they should state which ones are omitted from the script and why.
        \item At submission time, to preserve anonymity, the authors should release anonymized versions (if applicable).
        \item Providing as much information as possible in supplemental material (appended to the paper) is recommended, but including URLs to data and code is permitted.
    \end{itemize}

\item {\bf Experimental setting/details}
    \item[] Question: Does the paper specify all the training and test details (e.g., data splits, hyperparameters, how they were chosen, type of optimizer) necessary to understand the results?
    \item[] Answer: \answerYes{} 
    \item[] Justification: We have included the main details in the main text and rest in Appendix.
    \item[] Guidelines:
    \begin{itemize}
        \item The answer \answerNA{} means that the paper does not include experiments.
        \item The experimental setting should be presented in the core of the paper to a level of detail that is necessary to appreciate the results and make sense of them.
        \item The full details can be provided either with the code, in appendix, or as supplemental material.
    \end{itemize}

\item {\bf Experiment statistical significance}
    \item[] Question: Does the paper report error bars suitably and correctly defined or other appropriate information about the statistical significance of the experiments?
    \item[] Answer: \answerYes{} 
    \item[] Justification: We have included statistical tests for our model comparisons.
    \item[] Guidelines:
    \begin{itemize}
        \item The answer \answerNA{} means that the paper does not include experiments.
        \item The authors should answer \answerYes{} if the results are accompanied by error bars, confidence intervals, or statistical significance tests, at least for the experiments that support the main claims of the paper.
        \item The factors of variability that the error bars are capturing should be clearly stated (for example, train/test split, initialization, random drawing of some parameter, or overall run with given experimental conditions).
        \item The method for calculating the error bars should be explained (closed form formula, call to a library function, bootstrap, etc.)
        \item The assumptions made should be given (e.g., Normally distributed errors).
        \item It should be clear whether the error bar is the standard deviation or the standard error of the mean.
        \item It is OK to report 1-sigma error bars, but one should state it. The authors should preferably report a 2-sigma error bar than state that they have a 96\% CI, if the hypothesis of Normality of errors is not verified.
        \item For asymmetric distributions, the authors should be careful not to show in tables or figures symmetric error bars that would yield results that are out of range (e.g., negative error rates).
        \item If error bars are reported in tables or plots, the authors should explain in the text how they were calculated and reference the corresponding figures or tables in the text.
    \end{itemize}

\item {\bf Experiments compute resources}
    \item[] Question: For each experiment, does the paper provide sufficient information on the computer resources (type of compute workers, memory, time of execution) needed to reproduce the experiments?
    \item[] Answer: \answerYes{} 
    \item[] Justification: We have specified this in the Appendix~\ref{appx:cross-modal-implementation-details}. 
    \item[] Guidelines:
    \begin{itemize}
        \item The answer \answerNA{} means that the paper does not include experiments.
        \item The paper should indicate the type of compute workers CPU or GPU, internal cluster, or cloud provider, including relevant memory and storage.
        \item The paper should provide the amount of compute required for each of the individual experimental runs as well as estimate the total compute. 
        \item The paper should disclose whether the full research project required more compute than the experiments reported in the paper (e.g., preliminary or failed experiments that didn't make it into the paper). 
    \end{itemize}
    
\item {\bf Code of ethics}
    \item[] Question: Does the research conducted in the paper conform, in every respect, with the NeurIPS Code of Ethics \url{https://neurips.cc/public/EthicsGuidelines}?
    \item[] Answer: \answerYes{} 
    \item[] Justification: Yes, the research conducted in the paper conforms in every respect
    \item[] Guidelines:
    \begin{itemize}
        \item The answer \answerNA{} means that the authors have not reviewed the NeurIPS Code of Ethics.
        \item If the authors answer \answerNo, they should explain the special circumstances that require a deviation from the Code of Ethics.
        \item The authors should make sure to preserve anonymity (e.g., if there is a special consideration due to laws or regulations in their jurisdiction).
    \end{itemize}

\item {\bf Broader impacts}
    \item[] Question: Does the paper discuss both potential positive societal impacts and negative societal impacts of the work performed?
    \item[] Answer: \answerYes{} 
    \item[] Justification: This is mentioned in the Limitations Section~\ref{sec:limitation}
    \item[] Guidelines:
    \begin{itemize}
        \item The answer \answerNA{} means that there is no societal impact of the work performed.
        \item If the authors answer \answerNA{} or \answerNo, they should explain why their work has no societal impact or why the paper does not address societal impact.
        \item Examples of negative societal impacts include potential malicious or unintended uses (e.g., disinformation, generating fake profiles, surveillance), fairness considerations (e.g., deployment of technologies that could make decisions that unfairly impact specific groups), privacy considerations, and security considerations.
        \item The conference expects that many papers will be foundational research and not tied to particular applications, let alone deployments. However, if there is a direct path to any negative applications, the authors should point it out. For example, it is legitimate to point out that an improvement in the quality of generative models could be used to generate Deepfakes for disinformation. On the other hand, it is not needed to point out that a generic algorithm for optimizing neural networks could enable people to train models that generate Deepfakes faster.
        \item The authors should consider possible harms that could arise when the technology is being used as intended and functioning correctly, harms that could arise when the technology is being used as intended but gives incorrect results, and harms following from (intentional or unintentional) misuse of the technology.
        \item If there are negative societal impacts, the authors could also discuss possible mitigation strategies (e.g., gated release of models, providing defenses in addition to attacks, mechanisms for monitoring misuse, mechanisms to monitor how a system learns from feedback over time, improving the efficiency and accessibility of ML).
    \end{itemize}
    
\item {\bf Safeguards}
    \item[] Question: Does the paper describe safeguards that have been put in place for responsible release of data or models that have a high risk for misuse (e.g., pre-trained language models, image generators, or scraped datasets)?
    \item[] Answer: \answerNA{} 
    \item[] Justification: This paper poses no such risk.
    \item[] Guidelines:
    \begin{itemize}
        \item The answer \answerNA{} means that the paper poses no such risks.
        \item Released models that have a high risk for misuse or dual-use should be released with necessary safeguards to allow for controlled use of the model, for example by requiring that users adhere to usage guidelines or restrictions to access the model or implementing safety filters. 
        \item Datasets that have been scraped from the Internet could pose safety risks. The authors should describe how they avoided releasing unsafe images.
        \item We recognize that providing effective safeguards is challenging, and many papers do not require this, but we encourage authors to take this into account and make a best faith effort.
    \end{itemize}

\item {\bf Licenses for existing assets}
    \item[] Question: Are the creators or original owners of assets (e.g., code, data, models), used in the paper, properly credited and are the license and terms of use explicitly mentioned and properly respected?
    \item[] Answer: \answerYes{} 
    \item[] Justification: We cite the original paper from which the dataset in this paper was created.
    \item[] Guidelines:
    \begin{itemize}
        \item The answer \answerNA{} means that the paper does not use existing assets.
        \item The authors should cite the original paper that produced the code package or dataset.
        \item The authors should state which version of the asset is used and, if possible, include a URL.
        \item The name of the license (e.g., CC-BY 4.0) should be included for each asset.
        \item For scraped data from a particular source (e.g., website), the copyright and terms of service of that source should be provided.
        \item If assets are released, the license, copyright information, and terms of use in the package should be provided. For popular datasets, \url{paperswithcode.com/datasets} has curated licenses for some datasets. Their licensing guide can help determine the license of a dataset.
        \item For existing datasets that are re-packaged, both the original license and the license of the derived asset (if it has changed) should be provided.
        \item If this information is not available online, the authors are encouraged to reach out to the asset's creators.
    \end{itemize}

\item {\bf New assets}
    \item[] Question: Are new assets introduced in the paper well documented and is the documentation provided alongside the assets?
    \item[] Answer: \answerYes{} 
    \item[] Justification: We have documented the details of the evaluation dataset created in this paper in depth in \ref{sec:quality_evaluation}. The dataset is released with a dataset card describing its sources, fields, annotation process and license at \url{https://huggingface.co/datasets/aboots/SEMANTIX}.
    \item[] Guidelines:
    \begin{itemize}
        \item The answer \answerNA{} means that the paper does not release new assets.
        \item Researchers should communicate the details of the dataset\slash code\slash model as part of their submissions via structured templates. This includes details about training, license, limitations, etc. 
        \item The paper should discuss whether and how consent was obtained from people whose asset is used.
        \item At submission time, remember to anonymize your assets (if applicable). You can either create an anonymized URL or include an anonymized zip file.
    \end{itemize}

\item {\bf Crowdsourcing and research with human subjects}
    \item[] Question: For crowdsourcing experiments and research with human subjects, does the paper include the full text of instructions given to participants and screenshots, if applicable, as well as details about compensation (if any)? 
    \item[] Answer: \answerYes{} 
    \item[] Justification: This is reported in Appendix~\ref{appx:userstudy}.
    \item[] Guidelines:
    \begin{itemize}
        \item The answer \answerNA{} means that the paper does not involve crowdsourcing nor research with human subjects.
        \item Including this information in the supplemental material is fine, but if the main contribution of the paper involves human subjects, then as much detail as possible should be included in the main paper. 
        \item According to the NeurIPS Code of Ethics, workers involved in data collection, curation, or other labor should be paid at least the minimum wage in the country of the data collector. 
    \end{itemize}

\item {\bf Institutional review board (IRB) approvals or equivalent for research with human subjects}
    \item[] Question: Does the paper describe potential risks incurred by study participants, whether such risks were disclosed to the subjects, and whether Institutional Review Board (IRB) approvals (or an equivalent approval/review based on the requirements of your country or institution) were obtained?
    \item[] Answer: \answerYes{} 
    \item[] Justification: Study details in Appendix~\ref{appx:userstudy}.
    \item[] Guidelines:
    \begin{itemize}
        \item The answer \answerNA{} means that the paper does not involve crowdsourcing nor research with human subjects.
        \item Depending on the country in which research is conducted, IRB approval (or equivalent) may be required for any human subjects research. If you obtained IRB approval, you should clearly state this in the paper. 
        \item We recognize that the procedures for this may vary significantly between institutions and locations, and we expect authors to adhere to the NeurIPS Code of Ethics and the guidelines for their institution. 
        \item For initial submissions, do not include any information that would break anonymity (if applicable), such as the institution conducting the review.
    \end{itemize}

\item {\bf Declaration of LLM usage}
    \item[] Question: Does the paper describe the usage of LLMs if it is an important, original, or non-standard component of the core methods in this research? Note that if the LLM is used only for writing, editing, or formatting purposes and does \emph{not} impact the core methodology, scientific rigor, or originality of the research, declaration is not required.
    \item[] Answer: \answerYes{} 
    \item[] Justification:  We use GPT-5.4 as part of the semantic motion anchor generation pipeline, where it performs structured four-stage reasoning (handedness, motion, intent, verification) to compose physical motion narratives with speech transcripts into semantic motion anchors used as auxiliary contrastive supervision during training and also as LLM-as-a-judge.
    \item[] Guidelines:
    \begin{itemize}
        \item The answer \answerNA{} means that the core method development in this research does not involve LLMs as any important, original, or non-standard components.
        \item Please refer to our LLM policy in the NeurIPS handbook for what should or should not be described.
    \end{itemize}

\end{enumerate}

\end{document}